\documentclass[acmsmall,screen]{acmart}

\AtBeginDocument{%
  \providecommand\BibTeX{{%
    \normalfont B\kern-0.5em{\scshape i\kern-0.25em b}\kern-0.8em\TeX}}}

\usepackage{soul}
\usepackage{url}
\usepackage[utf8]{inputenc}
\usepackage{caption}
\usepackage{graphicx}
\usepackage{amsmath}
\usepackage{tabu}
\usepackage{amsthm}
\usepackage{booktabs}
\usepackage{algorithm}
\usepackage{algorithmic}
\usepackage{subfigure}
\usepackage{xspace}
\usepackage{xcolor}
\usepackage{multirow}
\usepackage{enumitem}
\usepackage{bm}
\usepackage{caption}

\usepackage{amsmath,amsfonts,amssymb}
\usepackage{listings}
\usepackage{makecell}
\usepackage{mathtools}
\usepackage{float}
\usepackage{footnote}
\usepackage[edges]{forest}
\newcommand{\paratitle}[1]{\textbf{#1}}
\newcommand{\ie}{\emph{i.e.,}\xspace}
\newcommand{\aka}{\emph{a.k.a.,}\xspace}
\newcommand{\eg}{\emph{e.g.,}\xspace}
\newcommand{\etal}{\emph{et~al.}\xspace}
\newcommand{\etc}{\emph{etc}}
\newcommand{\ignore}[1]{}

\newcommand{\tabincell}[2]{\begin{tabular}{@{}#1@{}}#2\end{tabular}}

\setcopyright{acmcopyright}

\begin{document}

\title{Pre-trained Language Models for Text Generation: A Survey}

\author{Junyi Li}
\authornote{Equal Contribution}
\affiliation{%
  \institution{Renmin University of China}
  \city{Beijing}
  \country{China}}
\affiliation{%
  \institution{Universit\'{e} de Montr\'{e}al}
  \city{Montr\'{e}al}
  \country{Canada}}
\email{lijunyi@ruc.edu.cn}

\author{Tianyi Tang}
\authornotemark[1]
\affiliation{%
  \institution{Renmin University of China}
  \city{Beijing}
  \country{China}}
\email{steventianyitang@outlook.com}

\author{Wayne Xin Zhao}
\authornote{Corresponding author}
\affiliation{%
  \institution{Renmin University of China}
  \city{Beijing}
  \country{China}}
\email{batmanfly@gmail.com}

\author{Jian-Yun Nie}
\affiliation{%
  \institution{Universit\'{e} de Montr\'{e}al}
  \city{Montr\'{e}al}
  \country{Canada}
}

\author{Ji-Rong Wen}
\affiliation{%
 \institution{Renmin University of China}
 \city{Beijing}
 \country{China}
 }


\begin{abstract}
  Text Generation aims to produce plausible and readable text in a human language from input data. The resurgence of deep learning has greatly advanced this field, in particular, with the help of neural generation models based on pre-trained language models (PLMs). Text generation based on PLMs is viewed as a promising approach in both academia and industry. In this paper, we provide a survey on the utilization of PLMs in text generation. We begin with introducing three key aspects of applying PLMs to text generation: 1) how to encode the input into representations preserving input semantics which can be fused into PLMs; 2) how to design an effective 
  PLM to serve as the generation model; and 3) how to effectively optimize PLMs given the reference text and to ensure that the generated texts satisfy special text properties. Then, we show the major challenges arisen in these aspects, as well as possible solutions for them. We also include a summary of various useful resources and typical text generation applications based on PLMs. Finally, we highlight the future research directions which will further improve these PLMs for text generation. This comprehensive survey is intended to help researchers interested in text generation problems to learn the core concepts, the main techniques and the latest developments in this area based on PLMs.
\end{abstract}

\begin{CCSXML}
<ccs2012>
   <concept>
       <concept_id>10002944.10011122.10002945</concept_id>
       <concept_desc>General and reference~Surveys and overviews</concept_desc>
       <concept_significance>500</concept_significance>
       </concept>
    <concept>
       <concept_id>10010147.10010178.10010179.10010182</concept_id>
       <concept_desc>Computing methodologies~Natural language generation</concept_desc>
       <concept_significance>500</concept_significance>
       </concept>
 </ccs2012>
\end{CCSXML}

\ccsdesc[500]{General and reference~Surveys and overviews}
\ccsdesc[500]{Computing methodologies~Natural language generation}
\keywords{Pre-trained Language Models, Natural Language Processing}

\maketitle

\section{Introduction}\label{sec:intro}

Text generation, also known as \emph{natural language generation}, has been one of the most important sub-fields in natural language processing (NLP). It aims to produce plausible and readable text in a human language, from the input data in various forms including text, image, table and knowledge base. %By processing the input data as semantic representations, text generation is desired to generate satisfactory output text.
%The underlying 
In the last decades, text generation techniques have been extensively applied to a wide range of applications. For example, they have been used in dialog systems to generate responses to user utterances in a conversation~\cite{xiaoice}, in machine translation to translate a text from one language into another~\cite{xlm}; and in text summarization to generate an abridged summary of the source text~\cite{summary_survey1}.

\begin{figure*}[t!]
    \centering
    \includegraphics[width=1\textwidth]{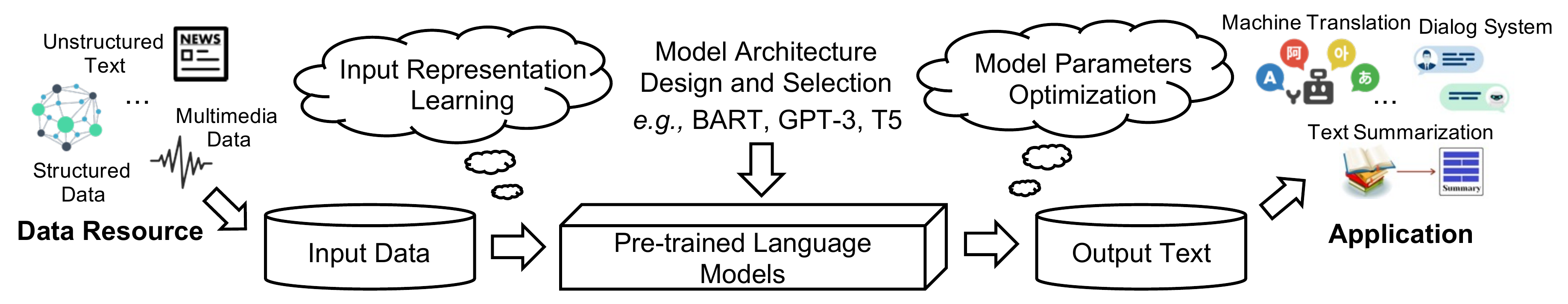}
    \caption{An illustrative process of applying PLMs to text generation. We divide the process into three main steps: input representation learning, model architecture design and selection, model parameters optimization.}
    \label{fig:example}
    \vspace{-0.4cm}
\end{figure*}

The primary goal of text generation is to automatically learn an input-to-output mapping from the data to construct an end-to-end solution with minimal human intervention. This mapping function allows the generation system to generalize in a broader field and to generate %more 
free text under the given conditions. %, which involve the learning of data representations  
Earlier approaches usually adopt statistical language models for modeling the conditional probabilities of words given an $n$-gram context~\cite{BrownCPPJLMR90,brown1995applying}. Such a statistical approach is known to suffer from the data sparsity issue, and a number of smoothing methods have been developed to alleviate this problem so as to better estimate unobserved term occurrences~\cite{ZhaiL01,TaoWMZ06}. Still, word tokens are used as the basic representation units in these approaches, which leads to the issue that similar tokens cannot be easily mapped with each other.

With the emergence of deep learning techniques~\cite{LeCunBH15}, neural network models have dominated the mainstream methods in text generation and make exceptional success in generating natural language texts. Deep neural generation models usually adopt the sequence-to-sequence framework~\cite{seq2seq} based on the encoder-decoder scheme: the encoder first maps the input sequence into fix-sized low-dimensional vectors (called \emph{input embeddings}), and then the decoder generates a target text based on the input embeddings. The representation by embeddings makes a key difference from earlier statistical approaches, which makes it easier to cope with the possible relations between inputs and outputs.
%Based on this framework, it becomes prevalent to develop end-to-end text generation systems. 
%\textcolor{blue}{
Various neural models have been proposed with different designs for the encoder-decoder architecture, such as graph neural networks (GNN) for encoding graph inputs~\cite{li2020knowledge} and recurrent neural networks (RNN) for decoding texts~\cite{li2019generating}. 
Besides, the attention mechanism~\cite{BahdanauCB14} and copy mechanism~\cite{pointer} are widely used to improve the performance of text generation models. An important merit of deep neural networks for text generation is that they enable  end-to-end learning of semantic mappings from the input data to output texts without labor-intensive feature engineering. Moreover, deep neural models employ low-dimensional 
semantic representations~\cite{iqbal2020survey} to capture linguistic features of language, which is  useful to alleviate data sparsity.
%}

%alleviate the \emph{feature engineering} problem, suffered in non-neural models. Non-neural models heavily rely on hand-craft features, while neural models usually use low-dimensional and dense vectors to implicitly represent linguistic features of text. 

Despite the success of deep neural models for text generation, a major performance bottleneck lies in the availability of large-scale labelled datasets. 
%The ability to learn effectively from raw text is crucial to alleviating the dependence on supervised learning in text generation. 
Most of text generation methods require substantial amounts of manually labelled parallel data, which restricts their applicability in many domains that suffer from a dearth of annotated examples. To date, most of existing labelled datasets for text generation tasks are usually small. In such cases, deep neural networks are likely to overfit on these small datasets and do not generalize well in practice. Moreover, the early neural models for text generation were still relatively shallow with only 1\textasciitilde3 neural layers. Therefore, these models have difficulties in modeling intricate relationships between the context and word meanings and deriving contextual word representations for better generation~\cite{abs-2003-08271}.

%Thus, in actually, the early neural models for many text generation tasks were still relatively shallow.
% the performance improvement may be less significant compared to other research fields (\eg computer vision). The main reason lies in that current datasets for most supervised text generation tasks are rather small (except machine translation). While, deep neural networks usually have a large number pf parameters, which make them overfit on these small training data and do not generalize well in practice. Thus, in actually, the early neural models for many text generation tasks were still relatively shallow.

In recent years, the paradigm of pre-trained language models (PLMs) is thriving in NLP~\cite{abs-2003-08271}. The basic idea is to first pre-train the models on large-scale unsupervised corpora and then fine-tune these models in downstream supervised tasks. Such a pretraining-finetuning framework achieved state-of-the-art performance.
With the emergence of Transformer~\cite{transformer} and higher computational power, the architecture of PLMs has evolved from shallow to deeper architectures, such as BERT~\cite{bert} and OpenAI GPT~\cite{gpt2}. Substantial work has shown that PLMs can encode massive amounts of linguistic knowledge from the pre-training corpora into their large-scale parameters and learn universal and contextual representations of the language with specially designed objectives such as masked token prediction. %language modeling. %These representations can be fine-tuned and transferred with little adaptation to a wide range of tasks using the corresponding supervised objective. 
Therefore, PLMs are generally beneficial for downstream tasks and can avoid training a new model from scratch. Following the success of PLMs in other NLP tasks, researchers have proposed to apply PLMs to text generation tasks with several steps (see Figure~\ref{fig:example})~\cite{gpt3,bart,t5}. Pre-trained on large-scale corpora, PLMs can understand natural language accurately and further express in human language fluently, both of which are critical abilities to fulfill text generation tasks. Grounding text generation on PLMs is seen as a promising direction in both academia and industry, which has much advanced the state of the art in this field. Thus, in this survey, we focus on text generation based on large PLMs. %, as this field has been totally transformed by these powerful PLMs.

There are a number of survey papers on text generation and on PLMs. For example, Qiu \etal~\cite{abs-2003-08271} summarized two generations of PLMs for the whole NLP domain and introduced various extensions and adaption approaches of PLMs. Kalyan \etal~\cite{ammus} gave a brief overview of the advances of self-supervised learning in Transformer-based PLMs. Han \etal~\cite{abs-2106-07139} took a deep look into the history of pre-training, especially its special relation with transfer learning and self-supervised learning. Besides, El-Kassas \etal~\cite{summary_survey1} %mainly paid attention to 
focused on the current application of PLMs to the field of text summarization. Zaib \etal~\cite{ZaibSZ20} discussed the application of PLMs to dialog systems with a special emphasis on question answering systems. These surveys %researches 
focused on specific applications, \eg summarization and dialogue systems, but did not go deep into the core technique, \ie text generation. As text generation is a key component in various applications, it is useful to provide a comprehensive survey on the topic of text generation based on PLMs. Differenbt from the existing surveys, this survey is intended to provide a more general description on this common task, rather than limiting it to a specific type of application.
It is worth noting that this survey is an extended version of the short survey~\cite{plm_survey}. The extensions include: (1) This paper covers a wider range of existing studies, %introduces more study efforts, 
evaluation protocols, open-source libraries, and common applications of PLMs-based text generation. % and a more concrete way, which 
This goes far beyond the scope of the previous short survey; (2) This paper provides a new schematic view involving %systematically defines 
three key aspects (\ie input data, model architecture, parameter optimization) about applying PLMs to text generation, which constitute the main content of this paper; (3) To provide a better picture of the existing solutions for various challenges, this paper includes more detailed descriptions and discussions about the %clear subsections to summarize 
their technical contributions.

%To the best of our knowledge, our survey is the first work that presents a comprehensive review of PLMs-based text generation. It aims to provide text generation researchers a synthesis and pointer to related researches. %Our survey also discusses some of the challenges and future directions.
The remainder of this survey is organized as follows. We first present the task formulation and an overview of PLMs in Section~\ref{sec:back}. Given the encoded input data, the goal of text generation is to optimize the generation function (\ie PLMs) for generating satisfactory output text. Thus, three key points are involved when applying PLMs to text generation: 1) how to encode the input data into representations preserving input semantics which can be fused into PLMs (Section~\ref{sec:input}); 2) how to design an effective %and performant 
PLM to serve as the generation function (Section~\ref{sec:arch}); and 3) how to optimize PLMs given the reference text and to ensure that the generated texts satisfy special text properties (Section~\ref{sec:optim}). Then, we discuss %figure out 
several typical non-trivial challenges and solutions within each key point in Section~\ref{sec:chal}. We present a summary of various useful resources to work with PLMs in Section~\ref{sec:eva} and common applications in Section~\ref{sec:app}. Finally, we %conclude and 
summarize the contribution of this survey and describe future directions in Section~\ref{sec:con}.

\section{Preliminary}\label{sec:back}

In this section, we first give a general task definition of text generation, then describe the background of PLMs, and finally introduce the three key aspects on PLM-based text generation methods.

\subsection{Text Generation}
Generally, a text can be modeled as a sequence of tokens $y=\langle y_1,...,y_j,...,y_n \rangle$, where each token $y_j$ is drawn from a vocabulary $\mathcal{V}$. 
The task of text generation aims to generate plausible and readable text in a human language. In most cases, text generation is conditioned on some input data (\eg  text, image, tabular, and knowledge base), which is denoted as $x$. In particular, the generated text is expected to satisfy some desired language properties such as fluency, naturalness, and coherence. We denote the desired properties for output text as a property set $\mathbb{P}$. 
Based on the above notations, the task of text generation can be formally described as:
\begin{equation}\label{eq:text_generation}
	y = f_\mathcal{M}(x, \mathbb{P}),
\end{equation}
where the text generation model $f_\mathcal{M}$ produces the output text $y$ given the input data $x$, satisfying some special proprieties from  the property set $\mathbb{P}$.
%$\bm{x}$ is the representation of the input data $x$, and $f_\mathcal{M}$ denotes the generation function.
In this survey, the text generation model $f_\mathcal{M}$ is specially crafted based on a PLM $\mathcal{M}$.

Specifically, according to the type of the input data $x$ and the property set $\mathbb{P}$, text generation can be instantiated into different kinds of tasks:

	$\bullet$~When the input data $x$ is \textbf{not provided} or is \textbf{a random vector}, text generation will degenerate into language modeling or unconditional text generation~\cite{gpt,gpt2}. In this case, the output text is required to satisfy some common language properties, such as fluency and naturalness.
	
	$\bullet$~When the input data $x$ is a set of \textbf{discrete attributes} (\eg topic words and sentiment labels), it becomes topic-to-text generation~\cite{pplm} or attribute-based generation~\cite{ctrl}. The input data $x$ plays the role of controlling the content %meaning 
	of the generated text. In such a situation, the output text should be relevant to the input topics or adhere to the required attributes. 
	
	$\bullet$~When the input data $x$ is \textbf{structured data} such as knowledge base or table, it is  considered as data-to-text generation~\cite{LiTZWYW21,tablegpt}. This task aims to generate a descriptive text about the structured data. Therefore, the output text should be objective and accurate.

    $\bullet$~When the input data $x$ is \textbf{multimedia input} such as image and speech, it becomes image captioning~\cite{abs-2003-01473} or speech recognition~\cite{fan2019unsupervised}. We may expect that the caption text %should 
    be lively for attracting children's attention, and the converted speech text be faithful to the original speech.

	$\bullet$~The most common form of input data $x$ is \textbf{a text sequence}. This form spans a number of applications such as machine translation~\cite{xlm}, text summarization~\cite{RotheNS20} and dialog system~\cite{dialogpt}. For a specific task, the output text is expected to satisfy desired properties. For example, the summaries in text summarization should not contradict the facts described in the input text, and the responses in dialog should be relevant to the input dialog history and context. %\textcolor{blue}{add another example}. %In machine translation, the order of semantic units (word, phrase, etc.) in both input and output text should be consistent. While in text summarization, the content in generated text should not contradict the facts in input text.

\subsection{Pre-trained Language Models}

Pre-trained language models (PLMs) are deep neural networks that are pre-trained on large-scale
unlabelled corpora, which can be further fine-tuned on various downstream tasks. It has been shown that PLMs can encode a significant amount of linguistic knowledge into their vast amounts of parameters~\cite{Ribeiro2020,plm_survey}. Therefore, it is promising to apply PLMs to enhance the understanding of language and improve the generation quality. %The idea of pretraining is inspired by human beings, \ie we transfer and reuse our old knowledge of what we have learned in the past to understand new knowledge and handle a variety of new tasks. 

Owing to the great success of Transformer~\cite{transformer}, almost all PLMs employ it as the backbone. As two typical PLMs,  GPT~\cite{gpt} and BERT~\cite{bert} are first built upon Transformer decoder and encoder respectively. Following GPT and BERT, PLMs such as XLNet~\cite{xlnet}, RoBERTa~\cite{roberta}, ERNIE~\cite{ernie}, T5~\cite{t5} and BART~\cite{bart} are propopsed in the literature. Among them, XLNet, RoBERTa and ERNIE are developed based on the BERT model, while T5 and BART are encoder-decoder based PLMs. Recent studies have shown that the performance of PLMs can be boosted by increasing the scale of model parameters~\cite{abs-2001-08361}, which triggered the development of large-scale PLMs such as GPT-3 (175B)~\cite{gpt3}, PANGU (200B)~\cite{pangu}, GShard (600B)~\cite{GShard} and Switch-Transformers (1.6T)~\cite{switch}, which consist of billions or trillions of parameters. In addition, %to language understanding and generation, 
PLMs are designed for other tasks such as named entity recognition~\cite{pires2019multilingual}, programming~\cite{feng2020codebert}, and networking~\cite{louis2020netbert}.
%Specifically, for text generation tasks, some of PLMs utilize the standard Transformer architecture following basic encoder-decoder framework, while others apply a decoder-only Transformer. 
According to the pre-training objectives, PLMs for text generation can be categorized as masked LMs, causal LMs, prefix LMs, and encoder-decoder LMs, which will be detailed in Section~\ref{sec:arch}.

\subsection{PLM-based Text Generation Methods}

To effectively leverage PLMs for downstream text generation tasks, we need to consider three key aspects from the perspectives of data, model, and optimization, respectively:

$\bullet$ \textbf{Input Data}: \emph{How to  encode the input $x$ into a representation preserving the input semantics that can be fused into the PLM $\mathcal{M}$?} For text generation, the input data, containing critical semantic information for the target output, often appears in various data types for different tasks (\eg sequential text, structured table, multimedia), whereas  most PLMs are typically pre-trained on the sequential text data. Therefore, it is a major challenge to develop effective, flexible representation learning approaches for PLMs to capture semantic information from various types of input data.

%For text generation, the input data, containing critical semantic information for the target output, is often not present in an appropriate form for PLMs, and the data forms also vary for different tasks. Therefore, we should develop effective, flexible representation learning approaches for capturing semantic evidence from the input in various data forms. 

%the input data contains important semantic information which is critical for generating the target text. Therefore, we need to develop effective representation learning approaches for capturing semantic evidence from the input.   Besides, the data forms of the input might be quite different across tasks, and we need to design effective encoders for different data forms of text generation tasks.
%To map the input data, the input data should be first transformed into representations to be fused with PLMs. Intuitively, the input data contains plenty of semantic information which is critical for generating satisfactory text. 

$\bullet$  \textbf{Model Architecture}: \emph{How to design an effective %and performant 
PLM $\mathcal{M}$ to serve as the generation function $f_\mathcal{M}$ and adapt to various text generation tasks?} In the literature, a number of PLMs have been developed with generalized architectures for general purposes %utility 
(\eg denoised auto-encoder~\cite{bart} or auto-regressive decoder~\cite{gpt2}),  While these general architectures cannot cope with some special text generation cases. Therefore, it is important to make specific designs on the underlying PLMs for achieving good task performance when adapting to different text generation tasks.

%In the literature, a number of PLMs have been developed with different architectures (\eg denoised autoencoder~\cite{bart} and autoregressive decoder~\cite{gpt2}).
%or pretraining tasks (\eg masked language models~\cite{bert} or text infilling~\cite{bart}).When adapting PLMs to text generation tasks, we need to make specific designs on the underlying PLMs in order to achieve good task performance. 
%carefully design the corresponding network architecture and pretraining tasks. 
%carefully design the network architecture and equip it with suitable pretraining tasks. 

%PLMs usually adopt various kinds of model settings (\eg model architecture, pretraining tasks), which have different effects on the downstream performance of PLMs. For example, BART adopts the denoising pretraining objectives, while GPT is pretrained with language modeling. Compared with GPT model, BART are more suitable for conditional text generation. 

$\bullet$ \textbf{Optimization Algorithm}: \emph{How to  optimize the text generation function (i.e., PLMs) $f_\mathcal{M}$ given the reference text $y$ and ensure that the generated text satisfies special text properties $\mathbb{P}$?} In order to produce satisfactory text, it is critical to learn the text generation function by developing effective optimization algorithms. A major challenge stems from the fact that some desired properties for output text are difficult to be formulated or optimized.

%In order to produce satisfactory text, it is key to learn the text generation function by developing effective optimization algorithms.

%After encoding the inputs and designing appropriate PLMs, it is important to develop effective optimization algorithms for producing satisfactory text.
%A major challenge lies in that  some desired properties for output text are difficult to be formulated or optimized. 

%\textcolor{blue}{Note: the three aspects should be written in a similar way: you raised a major challenge for the third aspect, but not for the other aspects. Overall, the three aspects are written in a vague way, it is not clear why the three aspects and how each aspect is important. The  italic questions are OK, but the explainations or clarifications that follow them are badly framed. }

%the final step is to optimize PLMs for generating output text. Besides, different tasks might require the generated text to satisfy various text properties. For example, the response text in dialogue systems should be relevant to the input dialog history and context.

\begin{figure*}
	\tiny
	\begin{forest}
		for tree={
			forked edges,
			grow'=0,
			draw,
			rounded corners,
			node options={align=center},
			calign=edge midpoint,
		},
		[PLMs for Text Generation, text width=1.3cm, fill=black!10
			[Encoding Input	Representations, text width=1.8cm, for tree={fill=red!20}
				[Unstructured, text width=1.1cm, for tree={fill=red!30}
					[Paragraph, text width=1.2cm
						[
						Hierarchy-based RL: DialogBERT~\cite{GuYH21}; Li~\etal~\cite{LiZF0Z20}; \\
						Graph-based RL: BASS~\cite{WuLXLCL0020}; Ouyang~\etal~\cite{OuyangZZ21},
						text width=6.9cm, node options={align=left}
						]
					]
					[Document, text width=1.2cm
						[
						Modeling Inter-Sentential Semantics: BERTSumExt~\cite{LiuL19}; HIBERT~\etal~\cite{hibert}; \\
						Capturing Critical Semantics: Nguyen~\etal~\cite{NguyenLLQ21}; Liu~\etal~\cite{LiuZZCDYW21}; \\
						RL Efficiency: Huang~\etal~\cite{HuangCPJW21}; DANCER~\cite{GidiotisT20}; Manakul~\etal~\cite{ManakulG20},
						text width=6.9cm, node options={align=left}
						]
					]
					[Multi-language, text width=1.2cm
						[
						Cross-Lingual: XLM~\cite{xlm}; CMLM~\etal~\cite{RenWLZM19}; \\
						Multi-Lingual: mBART~\cite{mbart}; mT5~\cite{mt5}; Wang~\etal~\cite{WangCZQL21},
						text width=6.9cm, node options={align=left}
						]
					]
				]
				[Structured, text width=1.1cm, for tree={fill=red!20}
					[Bridging Semantic Gap, text width=1.2cm
						[
						Structured Data Linearization: Ribeiro~\etal~\cite{Ribeiro2020}; TableGPT~\etal~\cite{tablegpt}; \\
						Representation Alignment: Li~\etal~\cite{LiTZWYW21},
						text width=6.7cm, node options={align=left}
						]
					]
					[Capturing Structural Information, text width=1.2cm
						[
						Incorporating Additional Objectives: TableGPT~\cite{tablegpt}; Mager~\etal~\cite{MagerANSLFR20}; \\
						Adding Structural Information as Input: Ribeiro~\etal~\cite{Ribeiro2020}; Fan~\etal~\cite{FanG20}; \\
						Employing Structural Encoding Module: StructAdapt~\cite{RibeiroZG21}; Li~\etal~\cite{LiTZWYW21},
						text width=6.7cm, node options={align=left}
						]
					]
					[
					Maintaining Text Fidelity, text width=1.2cm
						[
						Incorporating Additional Objectives: TableGPT~\cite{tablegpt}; Harkous~\etal~\cite{HarkousGS20}; \\
						Utilizing Copying Mechanism: Li~\etal~\cite{LiTZWYW21}; Suadaa~\etal~\cite{SuadaaKFOT20}; \\
						Adding Target Information as Input: Chen~\etal~\cite{ChenCZZZSW20},
						text width=6.7cm, node options={align=left}
						]
					]
				]
				[Multimedia, text width=1.1cm, for tree={fill=red!10}
					[Image Captioning, text width=1.2cm
						[
						XGPT~\cite{abs-2003-01473}; VisualGPT~\cite{chen2021visualgpt}; Yang~\etal~\cite{YangLW0FWZ0L21},
						text width=5.5cm, node options={align=left}
						]
					]
					[Video Captioning, text width=1.2cm
						[
						VideoBERT~\cite{SunMV0S19}; CBT~\cite{abs-1906-05743};  UniVL~\cite{luo2020univl},
						text width=5.5cm, node options={align=left}
						]
					]
					[Speech Recognition, text width=1.2cm
						[
						Fan~\etal~\cite{fan2019unsupervised}; Liao~\etal~\cite{LiaoSGSELQZ21},
						text width=5.5cm, node options={align=left}
						]
					]
				]
			]
			[Designing PLMs for Text Generation, text width=1.8cm, for tree={fill=green!20}
				[Standard Architecture, text width=1.1cm, for tree={fill=green!30}
					[Masked LM, text width=1.2cm
						[
						BERT2BERT~\cite{RotheNS20}; XLM~\cite{xlm},
						text width=6.0cm, node options={align=left}
						]
					]
					[Causal LM, text width=1.2cm
						[
						GPT-2~\cite{gpt2}; GPT-3~\cite{gpt3}; CPM~\cite{cpm}; CTRL~\cite{ctrl}; PanGu-$\alpha$~\cite{pangu},
						text width=6.0cm, node options={align=left}
						]
					]
					[Prefix LM, text width=1.2cm
						[
						UniLM~\cite{unilm}; UniLMv2~\cite{unilmv2}; GLM~\cite{glm},
						text width=6.0cm, node options={align=left}
						]
					]
					[Enc-Dec LM, text width=1.2cm
						[
						MASS~\cite{mass}; T5~\cite{t5}; BART~\cite{bart}; ProphetNet~\cite{prophetnet}; CPM-2~\cite{cpm2},
						text width=6.0cm, node options={align=left}
						]
					]
				]
				[Architecture Extensions, text width=1.1cm, for tree={fill=green!10}
					[Extended Embeddings, text width=1.2cm
						[
						Relative Position Embeddings~\cite{t5}; Hierarchical Position Embeddings~\cite{songnet}; 
						User Embeddings~\cite{plato}; Vowel Embeddings~\cite{deeprapper},
						text width=5.6cm, node options={align=left}
						]
					]
					[Improved Attention, text width=1.2cm
						[ 
						Multi-view Attention: Chen~\etal~\cite{ChenY20}; Liu~\etal~\cite{LiuZLRZY21}; \\
						Cross-attention: VECO~\cite{veco},
						text width=5.6cm, node options={align=left}
						]
					]
				]
			]
			[Optimizing PLMs for Text Generation, text width=1.8cm, for tree={fill=yellow!20}
				[Fine-Tuning for Text Generation, text width=1.1cm, for tree={fill=yellow!30}
					[Vanilla Fine-Tuning, text width=1.2cm
						[
						DialoGPT~\cite{dialogpt}; Ribeiro~\etal~\cite{Ribeiro2020},
						text width=6.0cm, node options={align=left}
						]
					]
					[Intermediate Fine-Tuning, text width=1.2cm
						[
						DAIFT: Liu~\etal~\cite{LiuWF21}; \\
						TAIFT: Fabbri~\etal~\cite{FabbriHLLGJRM21}; Mao~\etal~\cite{MaoMMC19},
						text width=6.0cm, node options={align=left}
						]
					]
					[Multi-Task Fine-Tuning, text width=1.2cm
						[
						Pure Multi-Task Fine-Tuning: Goodwin~\etal~\cite{GoodwinSD20}; Bai~\etal~\cite{Bai0H20}; \\
						Hybrid Multi-Task Fine-Tuning: Liu~\etal~\cite{LiuZZCDYW21};  Li~\etal~\cite{LiTZWYW21},
						text width=6.0cm, node options={align=left}
						]
					]
					[Parameter-Efficient Fine-Tuning, text width=1.2cm
						[
						Adapter-based Fine-Tuning: Houlsby~\etal~\cite{HoulsbyGJMLGAG19}; Ribeiro~\etal~\cite{RibeiroZG21}; \\
						Freezing-based Fine-Tuning: Gheini~\etal~\cite{gheini2021strengths}; \\
						Distillation-based Fine-Tuning: Chen~\etal~\cite{ChenGCLL20},
						text width=6.0cm, node options={align=left}
						]
					]
				]
				[Prompt-Tuning for Text Generation, text width=1.1cm, for tree={fill=yellow!20}
					[Discrete Prompts, text width=1.2cm
						[
						GPT-2~\cite{gpt2}; GPT-3~\cite{gpt3},
						text width=3.0cm, node options={align=left}
						]
					]
					[Continuous Prompts, text width=1.2cm
						[
						Prefix-Tuning~\cite{prefix-tuning}; Gu~\etal~\cite{abs-2111-02643},
						text width=3.0cm, node options={align=left}
						]
					]
				]
				[Property-Tuning for Text Generation, text width=1.1cm, for tree={fill=yellow!10}
					[Relevance, text width=1.2cm
						[
						TransferTransfo~\cite{transfertransfo}; DialoGPT~\cite{dialogpt}; Zeng~\etal~\cite{abs-2010-11140},
						text width=5.0cm, node options={align=left}
						]
					]
					[Faithfulness, text width=1.2cm
						[
						Kryscinski~\etal~\cite{KryscinskiPXS18}; TED~\cite{YangZGZ0D20},
						text width=5.0cm, node options={align=left}
						]
					]
					[Order-Preservation, text width=1.2cm
						[
						CSP~\cite{csp}; mRASP~\cite{mrasp}; Wada~\etal~\cite{abs-1809-02306},
						text width=5.0cm, node options={align=left}
						]
					]
				]
			]
		]
	\end{forest}
	\caption{The main content flow and categorization of this survey.}
    \label{fig:taxonomy}
    \vspace{-0.3cm}
\end{figure*}
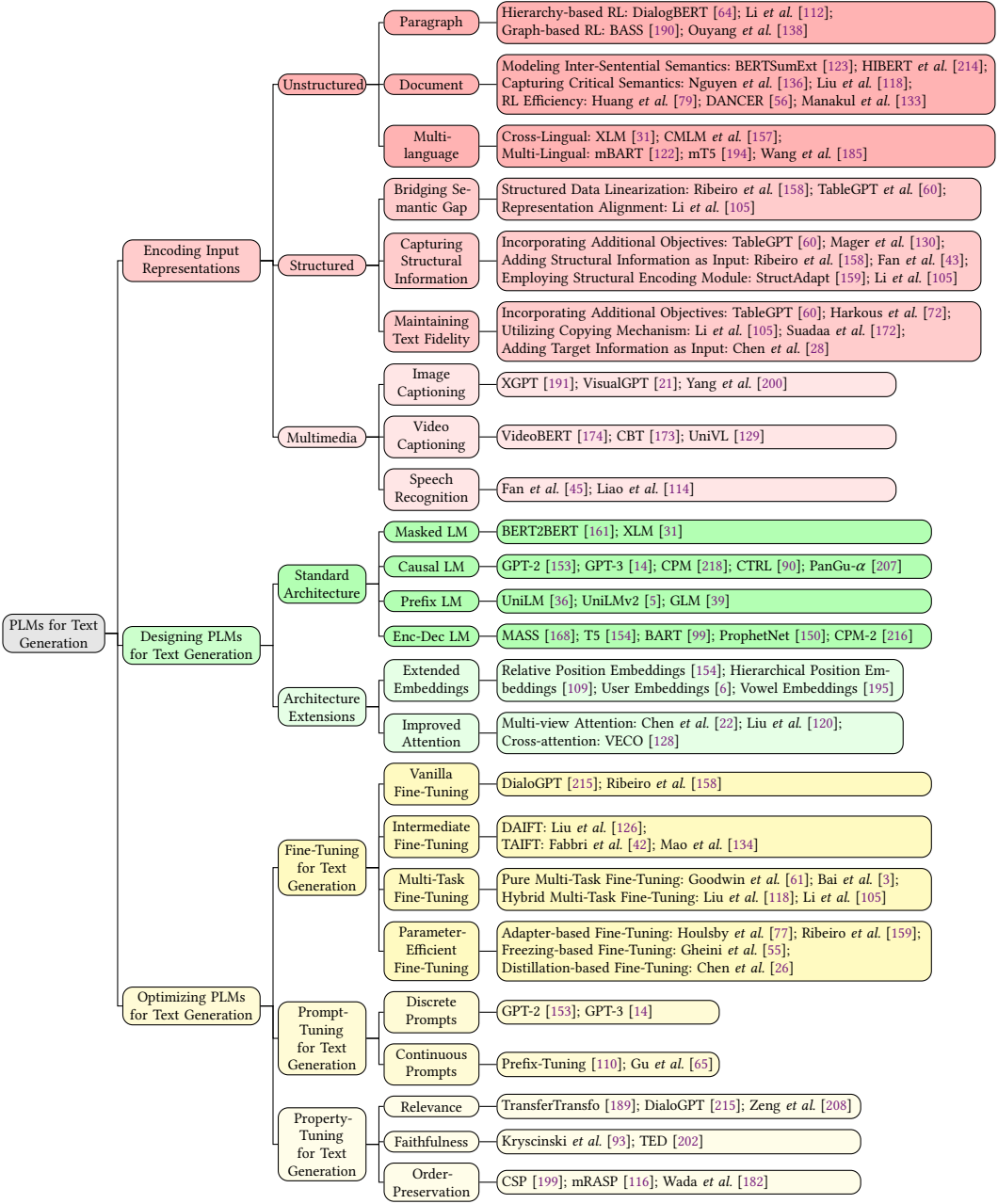

In the following sections, we will  present recent research efforts on PLM-based text generation, with an emphasis on the three aforementioned aspects. The overall organization of our description follows the schema shown in Figure~\ref{fig:taxonomy}.

\section{Encoding Input Representations}\label{sec:input}

As discussed in Section~\ref{sec:back}, the first aspect  is the encoding of %how to encode  
input data $x$ into meaningful representations preserving input semantics for PLMs. In this section, we will present three main types of input data for text generation, \ie unstructured input, structured input, and multimedia input.

\subsection{Unstructured Input}

In text generation, most studies focus on modeling unstructured text input (\eg sentence, paragraph, and document), which requires to accurately understand the input information and derive meaningful text representations. 
%requires an excellent capacity of language understanding beyond surface meaning of individual words in the input text. 
The aim of text representation learning is to condense the input text into low-dimensional vectors that can  preserve the core semantic meanings. In what follows, we will discuss how to derive effective semantic representations for three kinds of unstructured text data, namely paragraphs, documents and multi-lingual texts. 
%Edunov~\etal~\cite{EdunovBA19} adopted various strategies to integrate pretrained input text representations into sequence to sequence models, which verifies that pretrained representations are most effective when added to the generation model. 
%According to the semantic granularities of input text, we 
%Compared with traditional shallow models (\eg CNN~\cite{}), PLMs consist of large-scale parameters encoding rich semantic   knowledge, which is potentially beneficial to capture the core meaning of input text and helpful to generate accurate text. 

\subsubsection{Paragraph Representation Learning} A paragraph usually consists of multiple sentences describing different topics and each sentence contains a sequence of %several individual 
words. To capture both low-level word meanings and high-level topic semantics in a paragraph, many studies proposed hierarchy-based or graph-based methods to learn the paragraph representation.

\paratitle{Hierarchy-based Representation Learning.}~
%Due to the inclusion of different content, a paragraph can be viewed from hierarchical perspectives, resulting in multiple topic or discourse patterns. 
%For instance, based on what topics were discussed (topic view), a multi-turn conversation can be segmented into \textit{greetings, today’s weather, plan for today}; from a progression perspective (stage view), the same dialogue can be categorized into \textit{openings, intention} and \textit{discussion}~\cite{ChenY20}. 
For a multi-sentence paragraph such as a multi-turn dialogue, a typical approach is to concatenate sentences as a whole text and predict the output text~\cite{dialogpt,plato2}. However, flat concatenation cannot effectively capture the semantic dynamics across utterances which is likely to cause inaccurate generation. To deal with this issue, hierarchical encoders have been proposed to model the input paragraph~\cite{GuYH21, LiZF0Z20}. Gu~\etal~\cite{GuYH21} 
represented the dialogue context using DialogBERT, a hierarchical framework that utilizes sentence- and discourse-level Transformer encoders to encode each dialogue utterance and the sequence of utterance vectors, respectively. However, when encoding each individual utterance, it %lacks the consideration of 
does not consider the history information, which is essential for understanding dialogue utterances. Thus, Li~\etal~\cite{LiZF0Z20} employed a Transformer to encode each utterance into a dense vector, upon which a left-to-right flow module was designed to capture the utterance-level dynamic information flow.

%Chen~\etal~\cite{ChenY20} proposed to combine diverse views of the conversations, \ie structured views (topic view and stage view) and generic views (global view and discrete view). 

%Existing methods usually view a hierarchical text, \eg dialog text, as a linear sequence of tokens and learn to generate the next word through token-level self-attention. Such token-level encoding hinders the exploration of discourse-level coherence among utterances. Gu~\etal~\cite{GuYH21} presented a hierarchical architecture, DialogBERT, to efficiently capture the discourse-level coherence among utterances. In dialog systems, the flat pattern, \ie concatenating the dialogue history directly as the model input, will ignore the dynamic information flow across dialogue utterances. Li~\etal~\cite{LiZF0Z20} proposed a DialoFlow model, in which they introduce a dynamic flow mechanism to model the context flow, and design three training objectives to capture the information dynamics across dialogue utterances by addressing the semantic influence brought about by each utterance in large-scale pre-training.

\paratitle{Graph-based Representation Learning.} 
A long paragraph is likely to contain repeated, redundant or contradictory information.
%In a long paragraph, multiple sentences may contain %repeated, redundant or contradictory information. 
How to exploit the key semantics and remove minor information from the intricate paragraph text is critical to promote paragraph-based generation performance. Compared with sequences, by explicitly representing words or phrases as nodes and their relations (\eg similarity) as edges, graphs can easily aggregate relevant but disjoint context in the text~\cite{WuLXLCL0020,OuyangZZ21}. As a representative example, Wu~\etal~\cite{WuLXLCL0020} leveraged a phrase-level unified semantic graph, where nodes are phrases extracted by dependency parsing and relations are dependency %parsing 
relations. This graph can be used to aggregate co-referent phrases that are scattered in context for better capturing the long-range relations and global paragraph structures.
%Several graph augmentation methods can be designed to encode both the explicit and implicit relations in the text. 
Besides, in conversational machine reading, Ouyang~\etal~\cite{OuyangZZ21} formulated the input text as two complementary graphs, \ie explicit and implicit discourse graphs, to fully capture the discourse relations and latent vector interactions among all the elementary discourse units.

%These graphs can be used to encode the discourse relations for understanding local and contextualized connection within texts.

%Chen~\etal~\cite{ChenY21} proposed to explicitly model the rich structures in conversations by incorporating discourse relations between utterances and action triples (``who-doing-what'') in utterances through structured graphs to better encode conversations. Similarly, Wu~\etal~\cite{WuLXLCL0020} analyzed the long-distance relations in text based on a unified semantic graph, which aggregates co-referent phrases distributing across a long range of context and conveys rich relations between phrases. In particular, several graph augmentation methods can be designed to encode both the explicit and implicit relations in the text. Moreover, Ouyang~\etal~\cite{OuyangZZ21} proposed a dialogue graph modeling framework to improve the understanding and reasoning ability of machine. They proposed three types of textual graph to encode the input dialog from different perspectives. Specifically, Discourse Graph is designed to learn explicitly and encode the discourse relation among rule texts as well as the extra knowledge of scenario; Decoupling Graph is used for understanding local and contextualized connection within rule texts. And finally a global graph for fusing the encoded information together.

\subsubsection{Document Representation Learning} In many text generation tasks such as document translation and document summarization, the input text might be a long document consisting of multiple paragraphs. When encoding the documents, it is challenging to model cross-sentence (paragraph) semantics and capture the most critical semantics. 

\paratitle{Modeling Inter-Sentential Semantics.}~Most of PLMs are trained as masked language models. They mainly focus on learning token-level representations instead of sentence-level ones. Although segment embeddings are used to represent different sentences separately, they cannot capture the cross-sentence semantics. To encode inter-sentential semantics, several studies~\cite{LiuL19,ZhengL19,hibert} proposed to learn document representations in a hierarchical way. For example, Liu~\etal~\cite{LiuL19} inserted the ``[CLS]'' token at the beginning of each sentence to aggregate sentence-level features in lower layers and then combine them with self-attention in higher layers. Besides, Zhang~\etal~\cite{hibert} proposed HIBERT for learning  document representations in a hierarchical fashion by using a sentence encoder to map sentences into sentence vectors and a document encoder to further learn context-sensitive sentence representations given their surrounding sentence vectors as context.

\paratitle{Capturing Critical Semantics.}~In practice, sentences or paragraphs in long documents will inevitably complement, overlap, or conflict with one another. Therefore, it is necessary to retain the most critical contents and verbalize them in the generated text.
%Jin~\etal~\cite{JinWW20} has reported that excessively long input documents often lead to model degradation. 
%It is challenging for models to retain the most critical contents of complex input sequences, while generating the coherent, non-redundant, non-factual error and grammatically readable texts. Therefore, tasks involving long document input require models to have stronger capabilities for analyzing the corpora, identifying and representing consistent information. 
To address the issue of key points missing in output text, Nguyen~\etal~\cite{NguyenLLQ21} introduced a topic model to capture the global topic semantics of the document and a gate mechanism to control the amount of global semantics provided to the text generation module. 
%However, PLMs are originally pretrained on single sentence or short paragraphs, thus they are less capable of accurately modeling long-range dependencies and identifying key points in a long document. Considering this challenge, Zhang~\etal~\cite{hibert} proposed HIBERT, which stands for HIerachical Bidirectional Encoder Representations from Transformers, for learning the sentence and document representations in a hierarchical fashion. In particular, HIBERT uses a sentence encoder to transform each sentence into a vector and a document encoder to learn sentence representations given their surrounding sentences as context. 
%On the base of that, Xu~\etal~\cite{XuZWWZ20} introduced two pretraining tasks for HIBERT to obtain sentence-level self-attentions for rank-based extractive generation. 
Similarly, Liu~\etal~\cite{LiuZZCDYW21} proposed two topic-aware contrastive learning objectives, among which the coherence detection objective identifies topics of a dialogue by detecting the coherence change among topics and the sub-summary generation objective forces the model to capture the most salient information and generate a sub-summary for each topic.

%to capture global topic information and outline salient facts in a conversation. These objectives are able to implicitly model the topic shift \textcolor{blue}{varying upon conversations}, enforcing PLMs to focus more on snippets that contain salient information \textcolor{blue}{from the same topics} when generating output text.

%The former objective aims to implicitly model the topic change varying upon conversations, pushing PLMs to focus more on snippets that are more coherent and likely contain salient information from the same topics.

\paratitle{Representation Learning Efficiency.}~Efficiency is a crucial aspect for modeling long documents, especially when generating long text. Since the self-attention mechanism grows quadratically with sequence length, a number of studies aimed to improve the encoding efficiency of self-attention~\cite{HuangCPJW21,ManakulG20}. A representative example is  Manakul~\etal~\cite{ManakulG20}, which proposed local self-attention, allowing longer input spans during training; and explicit content selection, reducing memory and compute requirements. Furthermore, several researchers adopted divide-and-conquer encoding methods. By splitting the long document into short sentences, it is easier to summarize each short part of the document separately~\cite{GidiotisT20}, reducing the computational complexity.

%Huang~\etal~\cite{HuangCPJW21} proposed an efficient encoder-decoder attention with head-wise positional strides (HEPOS), where the attention heads follow a strided pattern and have varying starting positions. HEPOS reduces computational and memory costs while maintaining the power of emphasizing important tokens, and preserving the global context per head.
%Gidiotis~\etal~\cite{GidiotisT20} presented a divide-and-conquer method for encoding the complex and inter-sentential semantics of long documents. Specifically, they break a long document and its summary into multiple source-target pairs, which are used for training PLM that learns to summarize each part of the document separately. Compared with long documents, encoding semantics of short sentences will become simpler, reducing computational complexity and text noise. 
%To tackle the quadratic characteristic that self-attention mechanism grows quadratically with sequence length, Manakul~\etal~\cite{ManakulG20} proposed two methods: local self-attention, allowing longer input spans during training; and explicit content selection, reducing memory and compute requirements.

\subsubsection{Multi-lingual Representation Learning}

Existing PLMs are mainly pre-trained on English text while ignoring other low-resource languages. It is difficult to apply English-based PLMs to solve multi-lingual text generation tasks (\eg multi-lingual machine translation). Several approaches have been proposed to cope with multilingual texts.

\paratitle{Cross-lingual Representations.}~%Learning cross-lingual representations has been proven a beneficial method for cross-lingual transfer for some downstream tasks~\cite{KlementievTB12,ArtetxeLAC18,AhmadZMHCP19}. 
The core  idea of cross-lingual representation learning is to learn a shared embedding space for two languages, in order to improve PLMs' ability to translate between them. A well-known cross-lingual PLM is XLM~\cite{xlm}, which leveraged both monolingual and parallel data to learn cross-lingual representations. However, these learned representations on shared Byte-Pair Encoding (BPE) spaces is implicit and limited. Therefore, Ren~\etal~\cite{RenWLZM19} further computed cross-lingual $n$-gram embeddings and derived an $n$-gram translation table based on them for providing explicit representation learning signals.

\paratitle{Multi-lingual Representations.}~Given more than two languages, multi-lingual PLMs aim to learn representations for any of the languages. Based on English PLMs, BART and T5, Liu~\etal~\cite{mbart} and Xue~\etal~\cite{mt5} proposed mBART and mT5, respectively, which are pre-trained once for all languages. Considering the differences across languages (\eg syntactic rules), several studies utilized contrastive learning to learn multi-lingual representations~\cite{PanWWL20,WangCZQL21}. In particular, Wang~\etal~\cite{WangCZQL21} proposed two training objectives: contrastive sentence ranking (CSR) and sentence aligned substitution (SAS). CSR creates positive and negative sentence pairs based on their saliency scores, while SAS replaces sentences with those in another language. By contrastively learning these languages in a common text, the model can learn shared representation spaces across languages.

\subsection{Structured Input}

Structured data (\eg table, graph, and tree) is a critical kind of input for text generation in many real-world applications, such as medical report~\cite{HasanF19} and weather report~\cite{GoldbergDK94} generation. %Although PLMs can capture linguistic knowledge by pretraining on large-scale unlabeled dataset and generalize to downstream tasks in target domain, 
However, it is non-trivial to model structured input for PLMs due to three major challenges: (1) there exists a semantic gap between structured data and PLMs, since PLMs are typically pre-trained on natural language texts; (2) it is non-trivial to encode %lacks an encoding of the input structure which contains 
the structural information in the input data; (3) it requires to maintain fidelity of the generated text with respect to the input. % information when generating text.

\subsubsection{Bridging the Semantic Gap}

In general, PLMs are pre-trained on unstructured text, which differs in form from the structured data. % in semantic form. 
%To better leverage the structured data, we need to bridge the semantic gap between the structured input and PLMs' sequential nature.
Several methods have been proposed to bridge this gap.% between them.

\paratitle{Structured Data Linearization.}~In order to fit the structured input for PLMs, a simple approach is to linearize the input data into a \emph{sequence}~\cite{Ribeiro2020,MagerANSLFR20,FanG20}. Specifically, Ribeiro~\etal~\cite{Ribeiro2020} linearized knowledge graph (KG) into a sequence of triples by concatenating the relational triples. 
%Similaly, Mager~\etal~\cite{MagerANSLFR20} and Fan~\etal~\cite{FanG20} also linearized the AMR graph into a sequence of tokens. 
Besides, some studies adopted template-based heuristic methods to serialize the input data~\cite{tablegpt}. For example, the attribute-value pair ``\emph{name: james beattie}'' will be serialized as a sentence ``\emph{name is james beattie}''.
%In addition, Gong~\etal~\cite{tablegpt} employed a template-based method to serialize input table into text sequence. For example, the attribute-value pair ``\emph{name: jack reynolds}'' will be serialized as a sentence ``\emph{name is jack reynolds}''. Su~\etal~\cite{SuMBC21} utilized the ``prototypes'' to help the model bridging the structural gap between tables and texts. Specifically, a BERT-based prototype selector is required to select the prototypes as the input of PLMs, from the results retrieved by the IR system, that are closely related to the ta- ble for better guiding the neural generation model.

\paratitle{Representation Alignment.}~The semantic gap makes it difficult to effectively inject structured data representations into PLMs while directly serializing structured data. Therefore, some people proposed to align the structured data representations with PLM-based word embeddings in semantic spaces. For example, Li~\etal~\cite{LiTZWYW21} utilized graph neural networks (GNN) to project KG entities into embeddings, and then performed representation alignment by minimizing the Euclidean distance between the GNN-based and PLM-based entity embeddings.

\subsubsection{Capturing the Structural Information}

An important feature of structured data is that it represents data in a structural way,  such as the $\langle attribute, value \rangle$ pair in  table or the $\langle head, relation, tail \rangle$ triple in KB. Such structural information 
can be used to help generate faithful text by  modeling the input in a more accurate way. 

%can help models understand the input information correctly for generating more faithful text. 

%Unlike many text generation tasks that take sentences as input, data-to-text generation models need to process the input data with structural information.

\paratitle{Incorporating Additional Training Objectives.}~%Feeding PLMs with some sequential representation of the graph, such as a topological sorting, will loose some of the graph structure information. Complex graph annotations, such as AMR, also contain many special symbols and special constructs that departure from natural language and may by not interpretable by a PLM. 
%To explicitly model the structural information, 
To enhance the preservation of structural information, a typical approach is to  incorporate auxiliary training objectives related to structural information~\cite{tablegpt,LiTZWYW21,MagerANSLFR20}. One kind of objectives is to reconstruct the semantic structure of the input data. For example, Gong~\etal~\cite{tablegpt} utilized the  attribute names of input tables as the labels to reconstruct table structure based on the attribute value representations from PLMs, which enforces PLMs to embed table structure into table representations. Another method is to adjust the output text based on the structural information. Mager~\etal~\cite{MagerANSLFR20} proposed cycle-consistency based losses to assess the quality of output text based on how well it can reconstruct the input structure. 

%an alternative approach that combines PLMs with cycle consistency-based re-scoring, which aims at assessing the quality of a system's output based on how well an external 'reverse' system can reconstruct the input from it. Cycle-consistency based losses can be been used as part of the training objective to ensure PLMs to remember the graph structure.

\paratitle{Adding Structural Information as Input.}~As opposed to prior studies that implicitly capture structural information with training losses, several studies explicitly took structural information as input~\cite{Ribeiro2020,FanG20}. Ribeiro~\etal~\cite{Ribeiro2020} directly prepended ``⟨H⟩'', ``⟨R⟩'', and ``⟨T⟩'' tokens before the head entity, relation and tail entity of a KG triple to reveal the relations between entities. Besides, Fan~\etal~\cite{FanG20} used the graph embedding of an Abstract Meaning Representation (AMR) graph as input. The graph embedding provides the graph structure information by encoding the depth of each node (from the node to the root node) and the subgraph each node belongs to. %Such information allows the encoder to capture some graph structure information when modeling a sequence.

%They showed that, even though PLMs do not exhibit any graph-specific structural bias and KG is reduced to a mere bag of node and edge labels, PLMs still lead to remarkable generation performance. Rather than model the graph structure directly, 
%Fan~\etal~\cite{FanG20} modeled the AMR graph using a graph embedding. The graph embedding provides additional information to the encoder by encoding the depth of each node in the rooted graph and the subgraph each node belongs to. Concretely, each token has a word and position embedding, and additionally an indicator of depth calculated from the root and an indicator of which subtree the node belongs to. These additional embeddings are concatenated to the word and position embeddings. Such information allows the encoder to capture some graph structure information, while still modeling a sequence. Similarly, in addition to using the traditional token and position embeddings to encode the knowledge graph (KG), Wang~\etal~\cite{WangYLJR21} proposed a tree-level embedding method to capture the inter-dependency structures of the input graph.

\paratitle{Employing Structural Encoding Module.}~Since PLMs are originally developed for sequential input, it makes sense to incorporate additional modules to encode the structured input. A representative example is StructAdapt~\cite{RibeiroZG21}, which adds layer-wise graph convolution modules to learn representations built upon the graph connectivity over the PLM encoder. Similarly, Li~\etal~\cite{LiTZWYW21} employed GNN to  encode KG relations as embeddings, which will be taken as input of PLMs.

\subsubsection{Maintaining Text Fidelity}

In the literature of linguistics~\cite{Carroll16}, fidelity means the generated text adheres to the content in the structured data. Generating high-fidelity text that correctly describes the information of  structured input is the key to data-to-text generation algorithms. %In general, the output text should retain as much as important information from structured data. 

\paratitle{Incorporating Additional Training Objectives.}~To generate high-fidelity text adhereing to input, Gong~\etal~\cite{tablegpt} introduced an Optimal-Transport based content matching loss that measures the distance between the input information and the output text. Harkous~\etal~\cite{HarkousGS20} employed a semantic fidelity classification loss to detect and avoid generation errors such as hallucination.
%To explicitly learn graph-text alignments for improving output text's fidelity, Ke~\etal~\cite{KeJRCWSZH21} proposed three pre-training tasks including graph enhanced text reconstruction, text enhanced graph reconstruction, and graph-text embedding alignment to explicitly build the connection between knowledge graphs and text sequences.

\paratitle{Utilizing Copy Mechanism.}~The pointer-generator~\cite{pointer} is a typical method to ensure the faithfulness of generated text about input data by copying important words from input into output. For example, Li~\etal~\cite{LiTZWYW21} adopted pointer-generator to copy entities from input knowledge data to output text,  %Through grounding PLMs on external knowledge, it is likely to endow a generative PLMs with both rich knowledge and good generalization ability. 
and Suadaa~\etal~\cite{SuadaaKFOT20} copied table values into general placeholders to avoid producing hallucinated phrases that do not appear in the input table.

\paratitle{Adding Target Information as Input.}~To combat with the low-fidelity problem, Chen~\etal~\cite{ChenCZZZSW20} argued that it is important to leverage intermediate meaning representations to achieve faithful generation. Therefore, the authors enhanced the generation module with a logical form representing the semantics of the target text. %By separating logical reasoning and language realization, the correctness of the intermediate logical form is guaranteed, and the challenge for the realization module is fully shifted to semantic understanding.

%\paratitle{Employing Additional Module.}~Harkous~\etal~\cite{HarkousGS20} employed a weakly-supervised Semantic Fidelity Classifier (SFC) to detect and avoid generation errors (such as hallucination and omission). They also leveraged this classifier to assess outputs from any data-to-text system, overcoming the limitations of existing heuristic methods for detecting semantic errors.

\subsection{Multimedia Input}

In addition to the above textual data, 
multimedia data  (\eg image, video, and speech) has also been utilized as input of text generation algorithms, \eg image captioning and speech recognition.

%several attempts have been made to take the multimedia data as input (\eg image, video, and speech) such as image captioning and speech recognition. %Providing an effective representation of the input multimedia content is the first challenge of a multi-modal text generation pipeline.

\subsubsection{Image Captioning}

Image captioning, which aims to generate a textual description for an image, has been extensively studied in the field of computer vision (CV). Many studies have proposed multi-modal PLMs to combine textual and visual modalities. A well-known multi-modal PLM is XGPT~\cite{abs-2003-01473}. Inspired by text-based GPT, XGPT takes images as inputs and uses the image captioning task as the pre-training task in the pre-training stage. Chen~\etal~\cite{chen2021visualgpt} also proposed an image captioning PLM, called \emph{VisualGPT}. They designed a self-resurrecting attention mechanism to learn how to encode the visual information and adapt it to the PLM decoder. However, traditional vision-language pre-training fails to capture the relationship between the visual and text modalities. Yang~\etal~\cite{YangLW0FWZ0L21} proposed  three pre-training tasks to effectively learn better aligned representations among three kinds of input data: text word, visual object, and scene text.

%, which leverages the linguistic knowledge from PLMs. To balance between the prior linguistic knowledge of PLMs acquired from pretraining and the use of visual information in the image, they designed a self-resurrecting attention mechanism to learn how to encode the visual information and adapt it to PLMs decoder. In contrast to traditional vision-language pretraining that fails to capture scene text and its relationship with the visual and text modalities, Yang~\etal~\cite{YangLW0FWZ0L21} proposed to explicitly incorporate scene text during pretraining for downstream image caption tasks. With three pretraining tasks, including masked language modeling (MLM), image-text (contrastive) matching (ITM), and relative (spatial) position prediction (RPP), pre-training with scene text effectively helps the model learn a better aligned representation among the three modalities: text word, visual object, and scene text. 
%Inspired by generative pretraining objectives in GPT, Xia~\etal~\cite{abs-2003-01473} proposed a cross-modal pretrained model (XGPT) by taking images as inputs and using the image caption task as the basic generative task in the pretraining stage. To leverage underlying semantic and improve bidirectional generalization between two modalities, they designed three generative pre-training tasks, where the encoder takes single-stream or multi-stream data as input.

\subsubsection{Video Captioning}

Video captioning focuses on generating natural language text that can describe the video content. VideoBERT~\cite{SunMV0S19} and CBT~\cite{abs-1906-05743} are two early attempts to investigate video-language pre-training with regard to the video captioning task. %While, they performed pretraining only for the BERT-based encoder to learn bidirectional joint distributions over sequences of visual and linguistic tokens. So they have to train a separate video-to-text decoder, which tends to cause a \textit{pretrain-finetune discrepancy}. 
However, previous studies usually adopted one single encoder-decoder framework, which is not flexible for diverse downstream tasks. UniVL~\cite{luo2020univl} employed two single-modal encoders to encode text and video separately and a sentence decoder to generate video captions.
%Since they trained a separate video-to-text decoder, it tends to cause a \textit{pretrain-finetune discrepancy}. Therefore, UniVL~\cite{luo2020univl} encodes the text and video separately by two single-modal encoders, and then generates the text with a \textcolor{blue}{decoder (what decoder).}
%Following UniLM~\cite{unilm}, they pretrained the model on two masked language modeling (MLM) tasks, like cloze tasks designed for sequence-to-sequence LM. Luo~\etal~\cite{luo2020univl} proposed UniVL: a Unified Video and Language pre-training model for multimodal understanding and generation. 
%UniVL has four components, including two single-modal encoders, a cross encoder, and a decoder. The text and visual  are encoded separately by two single-modal encoders. Such a two-stream design is natural to retrieval tasks due to its scalability to very large datasets.

\subsubsection{Speech Recognition}

In practice, speech recognition is hungry for human-transcripted supervised data. Thus, a number of unsupervised and semi-supervised methods were developed to integrate PLMs for weakly-supervised learning. For example, Fan~\etal~\cite{fan2019unsupervised} proposed an unsupervised approach to pre-training encoder-decoder model with unpaired speech and transcripts. %Two pretraining stages are used to extract acoustic and linguistic information with speech and transcripts, which is useful for downstream speech recognition task. 
Liao~\etal~\cite{LiaoSGSELQZ21} proposed a speech recognition post-processing model that attempts to transform the incorrect and noisy recognition output into natural language text for humans and downstream tasks by leveraging the Metadata Extraction (MDE) corpus to construct a small task-specific dataset.

\section{Designing PLMs for Text Generation}\label{sec:arch}

After encoding the input data into low-dimensional representations, the next step is to develop an effective PLM $\mathcal{M}$ as the text generation function $f_{\mathcal{M}}$. %, formally denoted as $y=f_{\mathcal{M}}(\bm{x},\mathbb{P})$. Namely, we need to find a model $\mathcal{M}$, and minimize the probability $\text{Pr}_{\mathcal{M}}(y|\bm{x})$. Considering we aim to design a universe architecture, we dismiss the property set $\mathbb{P}$ for convenience.
Based on such an architecture of PLM, the text generation objective can be modeled as the conditional probability of the output text $y$ given the input data $x$, which can be formally factorized by tokens:
%the conditional probability is commonly modeled as an auto-regressive style, which can be factorized into:
\begin{equation}
	\text{Pr}_{\mathcal{M}}(y|x)=\prod_{i=1}^n\text{Pr}_{\mathcal{M}}(y_i|y_{<i}, x),
\end{equation}
where $y_i$ denotes the $i$-th output token, and $y_{<i}$ denotes the previous tokens $y_1,...,y_{i-1}$.

To compute the conditional probability, traditional neural models mainly adopt the RNN architecture~\cite{seq2seq} with several variants~\cite{pointer}. In recent years, solely based on attention mechanisms, Transformer~\cite{transformer} can better capture long-range dependency in texts, which is beneficial for modeling and generating texts. 
With the excellent parallelization capacities, Transformer has become the backbone for developing very large PLMs. 
When trained on large-scale unlabeled corpora~\cite{roberta}, PLMs built on the Transformer architecture can encode rich semantic or linguistic knowledge. Furthermore, it has been shown that PLMs can be effectively fine-tuned to different text generation tasks~\cite{bart,mass}. All these make PLMs the first choice to implement the text generation function $f_\mathcal{M}$.

\subsection{Standard Architecture}

Existing PLMs for text generation adopt either a single Transformer or a Transformer-based encoder-decoder as the backbone. PLMs, such as GPT-3~\cite{gpt3} and UniLM~\cite{unilm}, use a single Transformer encoder/decoder to simultaneously implement the process of input encoding and output decoding. This includes three major variants: masked LMs, causal LMs, and prefix LMs, with different attention mask strategies. In contrast, PLMs built upon Transformer encoder-decoder perform input encoding and output decoding separately. In the following, we describe these four  variants in detail. 

% Several PLMs follows the whole Transformer architecture, consisting of a stack of both encoder and decoder layers. While, PLMs can be pretrained using a single stack of encoder or decoder layers with different attention mask strategies. We divide PLMs into four categories, and discuss their details in the following sections.

\subsubsection{Masked Language Models}
Masked LMs use a full-attention Transformer encoder. Equipped with the full attention, models are usually pre-trained with masked language modeling~(MLM) task, \ie predicting the masked tokens using the bidirectional information. The most representative model is BERT~\cite{bert}, which is used extensively in natural language understanding~(NLU).

However, due to the discrepancy between the pre-training task of masked LMs and the downstream generation function, masked LMs are rarely utilized for text generation tasks~\cite{xlnet}. It is more common to use masked LMs as the encoder part for text generation, allowing to leverage the excellent bidirectional encoding capacities. For example, Rothe~\etal\cite{RotheNS20} proposed to initialize both the encoder and decoder of the generation model with BERT~\cite{bert}, which yields comparable performance  with other PLMs specially designed for text generation.

\subsubsection{Causal Language Models}

Similar to Transformer decoder, causal LMs adopt the diagonal mask matrix. Causal LMs are designed for language modeling, which is to determine the probability of a given sequence of words occurring in a sentence. Causal LMs are straightforward for text generation, predicting the next word conditioned on all previous words. 

In the literature, GPT~\cite{gpt} was the first causal LM for the text generation task. Then, GPT-2~\cite{gpt2} explored the transfer capacity of language models for zero-shot generation task, highlighting the significance of sufficient data. Furthermore, GPT-3~\cite{gpt3} showed that massive model parameters can significantly improve the downstream generation tasks, with a few examples or prompts.
%Although causal LMs show promising generation capabilities, it is not easy for them to control particular aspects of the generated text.
CTRL~\cite{ctrl} is proposed as a conditional causal LM to generate text based on control codes that govern style, content, and task-specific behavior. 
%CPM~\cite{cpm} and PanGu-$\alpha$~\cite{pangu} practiced on training large-scale auto-regressive Chinese language models. 
Causal LMs are simple and straightforward for text generation, but they have several structural and algorithmic limitations: Causal LMs encode the tokens just from left to right, thus ignore the bidirectional information on the input side. Moreover, causal LMs are not specially designed for the sequence-to-sequence generation tasks, thus in practice they do not achieve high performance in tasks such as summarization and translation~\cite{gpt2}.

\subsubsection{Prefix Language Models}

Upon a single Transformer, prefix LMs adopt bidirectional encoding scheme in the input side and natural left-to-right generation pattern in the output side.
%In order to overcome the disadvantages of the bidirectional masked LMs and the unidirectional causal LMs in text generation, Prefix LMs are proposed to combine the advantages of both. 
By utilizing the mixture attention mask, the tokens in the input text $x$ can attend to each other, while the tokens in the target text $y$ can only attend to all input tokens and previous generated tokens. 

UniLM~\cite{unilm} was the first prefix LM. Compared to causal LMs, UniLM used prefix attention mask to solve conditional generation tasks, similar to the encoder-decoder architecture. UniLMv2~\cite{unilmv2} and GLM~\cite{glm} improved vanilla prefix masking strategy by introducing permuted language modeling in XLNet~\cite{xlnet}. Although prefix LMs have several advantages, Raffel \etal~\cite{t5} compared single-Transformer prefix LMs to Transformer-based encoder-decoder LMs and concluded that adding explicit encoder-decoder attention is more effective to capture conditional dependencies.

\subsubsection{Encoder-Decoder Language Models}
Encoder-decoder LMs follow the standard Transformer architecture for text generation, consisting of stacks of both encoder and decoder layers. During pre-training, MASS~\cite{mass} and ProphetNet~\cite{prophetnet} took the sequence with one masked segment as the input of encoder and then the decoder generates the masked tokens in an auto-regressive way. T5~\cite{t5} randomly replaced several spans in the source text with different special tokens, and then  the decoder predicted 
every replaced span in turn. BART~\cite{bart} was pre-trained with denoising auto-encoder~(DAE), \ie the model learns to recover the original text from corrupted text, which is corrupted with different noising methods, such as sentence permutation and token deletion.

\subsection{Architecture Extensions}
To derive performant PLMs for text generation, many studies proposed to improve the Transformer backbone of PLMs. In this part, we will introduce two major improved techniques, \ie extended input embeddings and improved attention mechanism.

%, and we mainly focus on the attention modification in one Transformer layer.

\subsubsection{Extended Input Embeddings}
Besides (sub-)word embeddings, almost all PLMs use position embeddings to indicate the indices of input words. Compared to CNN and RNN, the self-attention operation is usually order-independent. Hence, it is essential to provide explicit position information to capture the sequential nature of text. Original Transformer~\cite{transformer} utilized the pre-determined absolute position embeddings with sinusoidal functions, while most PLMs (\eg BERT and GPT) adopted learned absolute position embeddings. Instead of absolute ones, relative position embeddings produce position embeddings according to the offset between two tokens. For example, T5~\cite{t5}, UniLMv2~\cite{unilmv2} and ProphetNet~\cite{prophetnet} employed an bucket relative positional method. In addition, hierarchical position embeddings are utilized to indicate inter- and intra- sentence position information, which is often used in some fixed-format text such as poem~\cite{songnet} and lyric~\cite{deeprapper}. 

Moreover, it is necessary to incorporate auxiliary embeddings to enrich the input information~\cite{ammus}. Similar to segment embeddings used in BERT, dialogue state embeddings~\cite{transfertransfo,plato} are used to assign each utterance, and user embeddings~\cite{plato,HamLJK20} are utilized to differentiate characters involved in a conversation. In the multilingual scenario, language embedding~\cite{mass,xnlg} is commonly introduced to inform the model about the language of each sentence. In addition, rhyme embeddings~\cite{songnet} and vowel embeddings~\cite{deeprapper} are proposed to indicate acoustics information in poem and lyric.
%Entity embeddings, triple embeddings and property embeddings~\cite{kgpt} reflect the information of knowledge graph.

\subsubsection{Improved Attention Mechanism} 
Although there exist various modules in Transformer (\eg position-wise FFN, self-attention, etc.), related works mainly focused on improving the self- and cross-attention mechanism for text generation~\cite{ammus}. In order to adapt to long-form text input and alleviate quadratic complexity of full-attention computation, sparse attention is proposed to replace the original self-attention for long-form input. Rather than attending to all other tokens, every token only attends to specific tokens with strategies such as window attention~\cite{bigbird,ManakulG20,PasunuruLBRD21}, global attention~\cite{bigbird,PasunuruLBRD21}, random attention~\cite{bigbird} and Sinkhorn attention~\cite{dialoglm}.

In practice, many text generation tasks need to process input data from multiple sources. It is common to leverage one or more encoders to encode multiple inputs. Therefore, several works proposed to utilize different strategies to aggregate multi-source inputs in the cross-attention module. 
Golovanov~\etal~\cite{GolovanovKNTTW19} conducted mean pooling for dialogue history, current state and persona information. Chen~\etal~\cite{ChenY20} and Liu~\etal~\cite{LiuZLRZY21} proposed multi-view attention and knowledge-aware attention to process embeddings from multiple views or knowledge sources. In addition, VECO~\cite{veco} pluged a cross-attention technique into the Transformer encoder to explicitly build the inter-dependence between multiple languages. BASS~\cite{WuLXLCL0020} and Ribeiro~\etal~\cite{RibeiroZG21} substituted the self-attention module with GNN to better extract structural information. Zeng~\etal~\cite{ZengN21} appended the gating mechanism after self-attention to inject condition-aware information.
\section{Optimizing PLMs for Text Generation}\label{sec:optim}

To obtain good performance, it is critical to develop effective optimization algorithms for PLM-based text generation models. 
%As discussed in Section~\ref{sec:input}, after the input data is encoded and the generation model (\ie PLMs) is designed, the next key step is to optimize PLM $\mathcal{M}$ for the text generation task. 
We consider three main types of optimization methods, namely fine-tuning, prompt-tuning, and property-tuning. We will detail each optimization method below.

%We consider that a PLM-based generation model can be optimized in three ways namely a) fine-tuning b) prompt-tuning and c) property-tuning. Next, we will describe each kind of way in detail. %Fine-tuning involves adapting model weights to downstream text generation tasks by minimizing task-specific losses. Prompt-tuning refers to formulating the text generation process as a pre-fix style generation which is close to the casual modeling objective. Property-tuning aims to optimize PLMs to generate text satisfying special language properties such as relevance and faithfulness.

\subsection{Fine-Tuning for Text Generation}

During pre-training, PLMs are able to capture general linguistic knowledge from large-scale corpora. However, it requires task-specific knowledge to perform downstream text generation tasks. For this purpose, fine-tuning is a popular approach to incorporating task-specific information into PLMs by adjusting their weights using downstream text generation datasets~\cite{gpt2}.

%Fine-tuning is able to impart task-specific knowledge to PLMs by adapting their weights with task-specific loss~\cite{bert}.
%, \ie for PLMs to perform well in downstream tasks, its weights should be close to the ideal setting to the target task~\cite{abs-1811-01088}. Fine-tuning imparts task-specific knowledge to PLMs by adapting its weights with task-specific loss~\cite{bert}. Moreover, fine-tuning enhances the model performance because it clusters the points of different labels away from each other such that there is a large separation between the cluster regions~\cite{abs-2106-14282}. Specifically, fine-tuning updates all the PLMs layers including the Transformer layers and the embedding layer but the higher layers are subjected to more changes compared to the lower layers~\cite{abs-2106-14282,MerchantRPT20,HaoDWX20}. In some cases where PLMs do not have unified input-output format across tasks, task-specific layers might also be added into PLMs during fine-tuning. 

According to how the parameters of PLMs are updated~\cite{ammus}, exiting fine-tuning methods for text generation can be categorized as 1) vanilla fine-tuning, 2) intermediate fine-tuning, 3) parameter-efficient fine-tuning, and 4) multi-task fine-tuning. Compared with vanilla fine-tuning, intermediate and multi-task fine-tuning can alleviate the overfitting issue on small text generation datasets to some extent. As the vanilla fine-tuning requires adjusting the entire model, parameter-efficient methods such as adapters~\cite{HoulsbyGJMLGAG19} can fine-tune PLMs in a lightweight manner.

\subsubsection{Vanilla Fine-Tuning}

Vanilla fine-tuning directly updates PLMs using downstream text generation datasets with task-specific losses (\eg cross-entropy loss~\cite{gpt2}).
%In vanilla fine-tuning, PLMs are directly applied to downstream text generation tasks based on task-specific losses such as the cross-entropy loss~\cite{gpt2}.
Zhang~\etal~\cite{dialogpt} trained the DialoGPT model on the basis of the GPT-2 architecture by modeling a multi-turn dialogue session as a long text and optimizing the generation model with language modeling objective. Ribeiro~\etal~\cite{Ribeiro2020} investigated two recent PLMs, BART and T5, for graph-to-text generation and fine-tuned them using the typical auto-regressive cross-entropy loss.
A major issue of vanilla fine-tuning is that it is often not sufficiently optimized on small datasets, which is prone to overfitting. 

%is prone to overfitting on small datasets. 

%Since it needs to re-adjust the large-scale parameters, 

%The disadvantage of vanilla fine-tuning is that PLMs with large-scale parameters are prone to overfit on small datasets. Moreover, with small datasets, the model weights cannot be optimized well which limits their performance on end tasks. %Intermediate fine-tuning or multi-task fine-tuning can avoid overfitting the model on small datasets.

\subsubsection{Intermediate Fine-Tuning}
\label{sec-ift}
The basic idea of intermediate fine-tuning is to incorporate an intermediate dataset consisting of sufficient labeled instances. The intermediate dataset can focus on the same target text generation task but from a different domain, or a similar NLP task from the same target domain. It is helpful to infuse domain- or task-specific knowledge from the intermediate dataset to alleviate the overfitting issue and enhance the performance on small target text generation datasets~\cite{abs-1811-01088}. According to the relatedness between the intermediate dataset and the target text generation dataset~\cite{ammus}, intermediate fine-tuning can be divided into two categories, \ie domain adaptive intermediate fine-tuning (DAIFT) and task adaptive intermediate fine-tuning (TAIFT).

%Intermediate fine-tuning involves fine-tuning PLMs on an intermediate dataset with a large amount of labelled instances. It helps PLMs to gain domain- or task-specific knowledge from the intermediate dataset which avoids overfitting and enhances their performances on small target datasets~\cite{abs-1811-01088}. According to the correlation between intermediate dataset and target dataset, intermediate fine-tuning can be divided into two categories, \ie domain adaptive intermediate fine-tuning (DAIFT) and task adaptive intermediate fine-tuning (TAIFT).

\paratitle{Domain Adaptive Intermediate Fine-Tuning.} According to Kalyan~\etal~\cite{ammus}, DAIFT 
utilizes an intermediate dataset, which focuses on a similar NLP task (not text generation tasks) from the same target domain, consisting of sufficient labeled instances. 
By leveraging such an intermediate dataset, PLMs can be enriched with domain-specific knowledge, which is helpful to improve the performance of the target text generation task within the same domain. 
%involves fine-tuning PLMs on the same domain intermediate dataset with a large number of labelled instances, \ie intermediate and target datasets are of the same domain but different tasks. DAIFT can impart more domain-specific knowledge to PLMs which enhances the model performance on the same domain target task. 
DAIFT is commonly used in machine translation to eliminate the issue of unseen languages in translation pairs. For example, to improve the translation quality of the low-resource target language (\eg Kazakh), Liu~\etal~\cite{LiuWF21} constructed a large-scale intermediate monolingual corpus of the target language and fine-tuned mBART by reconstructing the corrupted target-language text. The intermediate dataset comes from the same language domain as the target dataset (\eg Kazakh), which can impart language-related linguistic knowledge to PLMs for a better translation performance.

%. Here the intermediate dataset (monolingual corpus of the target language) and target dataset (target language pairs) are from the same domain. 
%Based on the same domain dataset, Xing~\etal~\cite{XingW21} proposed three self-supervised tasks to help PLMs understand the structured input table. The first two tasks are only used to train the encoder while keeping the decoder frozen, and the third task are used to train the encoder and decoder at the same time.

\paratitle{Task Adaptive Intermediate Fine-tuning.} In contrast with DAIFT, TAIFT incorporates an 
intermediate dataset on the same target text generation task but from a different domain. It aims to infuse task-specific knowledge from the massive intermediate labeled dataset for improving the same target text generation task. 
%On the other hand, TAIFT means fine-tuning PLMs on the same or related task dataset with a large amount of labelled examples, \ie intermediate and target datasets are focused on the same or related task. In this way, the intermediate and target datasets need not be from the same domain. Compared with DAIFT, TAIFT imparts more task-specific knowledge to PLMs which enhances the model performance on the same target task. %For example, Maurya~\etal~\cite{MauryaDKD21} firstly fine-tuned mBART~\cite{mbart} with an intermediate task using monolingual data of three languages. The objective function of the intermediate task is close to the target tasks which enriches the multi-lingual latent representation of mBART and provides good initialization for target tasks. 
It has been shown that the additional training with general-purpose text corpora (\eg  Wikipedia, WebText) on the same text generation task can improve the performance on a specific domain (\eg Movie)~\cite{FabbriHLLGJRM21,MaoMMC19}.
%completing the same target text generation task on intermediate general-purpose text corpora (\eg  Wikipedia, WebText) can improve the performance of the target text generation task on a specific domain (\eg Movie)~\cite{FabbriHLLGJRM21,MaoMMC19}.
%TAIFT on \textcolor{blue}{general domains (what are general domains)} can improve the performance of PLMs on the same specific task~\cite{FabbriHLLGJRM21,MaoMMC19} \textcolor{blue}{rewritting the entire sentence}. 
For example, Fabbri~\etal~\cite{FabbriHLLGJRM21} performed summarization on intermediate pseudo-summaries created from Wikipedia to improve the zero-shot and few-shot performance of abstractive summarization, and Mao~\etal~\cite{MaoMMC19} conducted generation on intermediate BookCorpus  dataset (built from WebText) to improve commonsense story generation on the target WritingPrompts dataset. %Furthermore, several researchers combined DAIFT and TAIFT in practice by dividing the parameters of PLMs into domain-related and task-related parameters and then using DAIFT and TAIFT to optimize them respectively~\cite{LiuZLRZY21}. 

%Similarly, Chen~\etal~\cite{ChenY21a} also improved semi-supervised summarization model performance by first pre-train PLMs on unlabelled conversations with pseudo summaries and then fine-tune it on labelled instances. %Yang~\etal~\cite{YangZGZ0D20} first leveraged the lead bias in news articles to pretrain PLMs on millions of unlabelled corpora, and then finetuned PLMs on target domains through theme modeling and a denoising autoencoder to enhance the quality of generated summaries. 
%Liu~\etal~\cite{LiuZLRZY21} divide the parameters of a PLM into dialogue-related and knowledge integration-related. They utilized supervised learning to pre-train dialogue-related parameters on general domain dialogues (\eg online forum comments), and perform domain-adaptive pre-training to initialize knowledge-related parameters. %Mao~\etal~\cite{MaoMMC19} performed intermediate fine-tuning of the pretrained GPT on BookCorpus as a method of domain adaptation from WebText to the domain of stories, and then fine-tuned on the target WritingPrompts dataset with a multi-task learning objective.

\subsubsection{Multi-Task Fine-Tuning} Multi-task fine-tuning can exploit cross-task knowledge to improve the primary text generation task by incorporating auxiliary tasks. 
%The primary aim of multi-task learning is to improve the performance of target tasks with the help of auxiliary tasks. Intuitively, multi-task learning can help PLMs learn useful knowledge across tasks.
%By learning from multiple diverse datasets simultaneously, multi-task learning reduces the requirement of a large number of labelled instances in the target task. According to the similarity between the primary task and auxiliary tasks, multi-task fine-tuning (MTFT) can be divided into two categories, \ie pure MTFT and hybrid MTFT.
Furthermore, by obtaining knowledge from related NLP tasks, multi-task fine-tuning can enhance the robustness of PLMs and reduce the need for large amounts of labeled instances in the  text generation task. According to the similarity between the primary text generation task and auxiliary tasks, multi-task fine-tuning (MTFT) can be divided into two categories, \ie pure MTFT and hybrid MTFT.

%multi-task learning reduces the requirement of a large number of labelled instances in the target task. According to the similarity between the primary task and auxiliary tasks, multi-task fine-tuning (MTFT) can be divided into two categories, \ie pure MTFT and hybrid MTFT.

\paratitle{Pure Multi-Task Fine-Tuning.} Pure MTFT incorporates auxiliary tasks that are the same as the primary text generation task but from different domains. Previous studies mainly utilized additional datasets to eliminate the data scarcity issue of the primary text generation task~\cite{GoodwinSD20,Bai0H20}. Specifically, Goodwin~\etal~\cite{GoodwinSD20} leveraged twenty-one additional summarization datasets to improve zero-shot summarization on previously unseen datasets. Besides, Bai~\etal~\cite{Bai0H20} incorporated an auxiliary monolingual summarization task to improve the primary cross-lingual summarization task in a low-resource language.
%To capture the relationships between the discrete phrases of summaries in different languages, they employed one unified decoder to generate the sequential concatenation of monolingual and cross-lingual summaries for establishing reliant connections between monolingual summarization and cross-lingual summarization tasks.

\paratitle{Hybrid Multi-Task Fine-Tuning.} Hybrid MTFT incorporates auxiliary tasks that are different from the primary text generation task.
These diverse auxiliary tasks can enhance the primary generation task in different aspects.
%Incorporating diverse auxiliary tasks is mainly focused on enhance special needs of the primary generation task. 
For example, Liu~\etal~\cite{LiuZZCDYW21} and Jin~\etal~\cite{JinJZOS20} fine-tuned PLMs with auxiliary tasks (\eg coherence detection, style-carrying text reconstruction) to control the content of the generated text according to the topic change and text style (humor, romance, and clickbait). Besides, to improve the faithfulness of the generated text, Li~\etal~\cite{LiTZWYW21} and Gong~\etal~\cite{tablegpt} introduced auxiliary input reconstruction tasks to reconstruct KG triples and table values for aligning the input information with the generated content.

\subsubsection{Parameter-Efficient Fine-Tuning}
%When applying PLMs to diverse downstream tasks, 
%Fine-tuning adapts PLMs' weights to downstream tasks by minimizing the task-specific loss, \ie fine-tuning starts with copying the entire model weights and making small changes. 
As the above fine-tuning methods require updating all PLM parameters, it is time-consuming to perform the entire fine-tuning in  resource-limited scenarios. Many studies developed parameter-efficient fine-tuning (PEFT) for text generation tasks. 

%difficult to perform such an full-parameter approach in  resource-limited scenarios.
%it needs to train a separate model for each task which is not parameter efficient.  involves updating the entire model weights, it is required to train a separate model for each task which is not parameter efficient. Therefore, fine-tuning PLMs in a parameter-efficient way is urgent in resource-limited scenarios. %Adapters~\cite{HoulsbyGJMLGAG19}, freeze-based fine-tuning and distillation-based fine-tuning helps to fine-tune the model in a parameter- efficient way.

\paratitle{Adapter-based Parameter-Efficient Fine-Tuning.} 
Adapter is a special neural layer proposed by Houlsby~\etal~\cite{HoulsbyGJMLGAG19} to fine-tune PLMs in a parameter-efficient way. %The adapter consists of two feed-forward layers with a non-linear layer in between and a skip connection. 
The adapter module projects the input vector into a small vector and then projects back into the original dimension using two feed-forward layers and a non-linear layer. Specifically, the adapters first project the original $d$-dimensional features into a smaller dimension, $m$, apply a non-linearity, then project back to $d$ dimensions. The total number of parameters added per layer, including biases, is $2 m d+d+m$. By setting $m \ll d$, we can limit the number of additional parameters per task. Thus, it is highly efficient to fix the parameters of original PLMs but only fine-tune the adapters~\cite{SticklandLG21,ChenS21a}.
%The bottleneck dimension, $m$, provides a simple means to trade-off performance with parameter efficiency. %The adapter module itself has a skip-connection internally. With the skip-connection, if the parameters of the projection layers are initialized to near-zero, the module is initialized to an approximate identity function. 
To address the inefficiency and overfitting issues in low-resource abstractive summarization,  Chen~\etal~\cite{ChenS21a} inserted adapters into both encoder and decoder of PLMs and only fine-tuned the adapters. A number of studies have shown that adapters can help PLMs efficiently capture some input characteristics for generating more accurate output text with a low extra cost in terms of parameters~\cite{LePWGSB20,RibeiroZG21}. For example, Ribeiro~\etal~\cite{RibeiroZG21} utilized adapters to model the input graph structure effectively when fine-tuning PLMs on graph input.

%, and Le~\etal~\cite{LePWGSB20} used adapters to specialize speech translation to specific language pairs. \textcolor{blue}{Can't see the relations to text generation. Need rewriting}

%method to encode graph structure into PLMs by modeling interactions among the nodes based on the graph connectivity, only training graph structure-aware adapter parameters. %Le~\etal~\cite{LePWGSB20} investigated adapters for multilingual speech translation, which shows that adapters can be used to efficiently specialize speech translation to specific language pairs with a low extra cost in terms of parameters.

\paratitle{Freezing-based Parameter-Efficient Fine-Tuning.} This approach refers to freezing most parameters and only updating a small proportion of PLM parameters. Recent studies have shown that not all parameters of PLMs are necessary to be fine-tuned for text generation tasks, and some of them can be fixed during fine-tuning without large impact on the model performance. 
Several studies also revealed that cross-attention (or encoder-decoder attention) layers are more important than self-attention layers when fine-tuning PLMs for machine translation~\cite{gheini2021strengths,YouSI20}. Therefore, Gheini~\etal~\cite{gheini2021strengths} only fine-tuned cross-attention layers while kept the encoder and decoder fixed. This approach  achieved comparable translation performance to fine-tuning all parameters. %Besides, Stickland~\etal~\cite{SticklandLG21} freezed most of the model parameters and added extra positional embeddings when fine-tuning BART on English monolingual data.  \textcolor{blue}{Can't see the relations to text generation. Need rewriting}

\paratitle{Distillation-based Parameter-Efficient Fine-Tuning.} Another parameter-efficient fine-tuning approach is to distill  large teacher PLMs into small student models. By distilling the knowledge in PLMs for text generation into small generative models (\eg LSTM), the student models can be efficiently fine-tuned for better generation performance~\cite{abs-2010-13002,ChenGCLL20}. As a representative example, Chen~\etal~\cite{ChenGCLL20} leveraged BERT as the teacher model that generates sequences of word probability logits and treated the Seq2Seq model as the student network, which can effectively learn from the teacher's outputs.

%Shleifer~\etal~\cite{abs-2010-13002} proposed a ``shrink and fine-tune'' approach that extracts a small student model from the maximally spaced layers of a fine-tuned teacher PLM. \textcolor{blue}{Can't see the relations to text generation. Need rewriting and extension}

\subsection{Prompt-Tuning for Text Generation}

Most generative PLMs are pre-trained using language modeling objectives and then fine-tuned on text generation tasks with task-specific objectives. Such a discrepancy between pre-training and fine-tuning affects the performance of PLMs on text generation tasks. As a new learning paradigm, prompt learning~\cite{prompt-survey}
reformulates the downstream  tasks (text generation tasks) into the language modeling task in pre-training.

%The downstream performance of PLMs can be improved by prompt-tuning especially in few-shot and zero-shot settings. Here the prompt can be close or pre-fix style and it can be generated manually or automatically~\cite{prompt-survey}. Specifically, close style prompts are suitable for PLMs pretrained by masked language modeling objective while pre-fix style prompts are suitable for PLMs using casual modeling objective. Therefore, pre-fix prompts are usually used in text generation, as they mesh well with the left-to-right nature of the model.

\subsubsection{Background}

According to Liu~\etal~\cite{prompt-survey}, a prompt function $f_{prompt}(\cdot)$ converts the input text $x$ into a prompt $x' = f_{prompt}(x)$ through a two-step process:

\begin{enumerate}
\item [1.] Apply a textual \textit{template} containing two slots: an input slot $[X]$ for input $x$ and an answer slot $[Z]$ for an intermediate generated answer text $z$ that will later be mapped into $y$.
\item [2.] Fill the input slot $[X]$ with the input text $x$.
\end{enumerate}

Here the prompt can be \emph{cloze} or \emph{prefix} style. The cloze-style prompt is usually adopted in language understanding tasks, where the empty slot $[Z]$ is either in the middle of the prompt or at the end. For example, in sentiment analysis where $x=$``\emph{I love this movie}'', the template may take a clozed form such as ``$[X]$ It was a really $[Z]$ movie.'' to predict the answer in $[Z]$. While in the prefix-style prompt, the input text comes entirely before the empty slot $[Z]$ such as ``English: $[X]$ German: $[Z]$'' in machine translation. Prefix prompts are widely used in text generation, as they mesh well with the left-to-right nature of language modeling. In the above prompt examples, the template is composed of \textit{discrete} natural language tokens, but the tokens can also be virtual words (\eg represented by numeric IDs), which would be mapped into \textit{continuous} embeddings later.

\subsubsection{Discrete Prompts}

Early prompting studies create prompts by manually designing templates based on human introspection. As a pioneering study, GPT-2~\cite{gpt2} performed text generation tasks using various manually-created prompts. For example, the prompt ``translate to french, [\textit{input}], [\textit{output}]'' is used in machine translation. The prompt defines the semantic mapping from input data to output text in a specific text generation task. By utilizing diverse prompts, a single PLM is able to perform a number of different text generation tasks. %Furthermore, Brown~\etal~\cite{gpt3} perform in-context learning in GPT-3 for text generation, creating a prompt with manual templates and augmenting the input with multiple answered prompts. 
%Schick and Sch\"{u}tze~\cite{abs-2012-11926} explore fixed-prompt LM tuning for few-shot text summarization with manually crafted templates. While, most of recent studies about discrete prompts are still far from sufficient for controllable text generation. It is common for a language model to deviate the generation process from the original prompt and start generating text of unrelated topics. To tackle this challenge, Zou~\etal~\cite{inverse-prompt} proposed a novel discrete prompting method, inverse prompting, to refine the process of text generation from PLMs. Given a piece of generated text, an inverse prompt is constructed using the generated text, and then the conditional likelihood score of the original prompt given the inverse prompt is computed based on PLMs. The best generation candidates are finally selected according to the conditional likelihood score. As a result, inverse prompting ensures that PLMs predict the prompt given the generated text with high likelihood, which encourages the relevance of the generated text to the prompt. 
These approaches heavily relied on manual efforts to create prompts; but PLMs are highly sensitive to prompts: improperly-created prompts lead to low performance~\cite{JiangXAN20}. To avoid the need to manually specify prompts, Shin~\etal~\cite{autoprompt} proposed AutoPrompt to automatically search for template tokens. Several other methods have also been proposed to discover discrete prompts automatically such as paraphrasing existing prompts~\cite{JiangXAN20}, generating prompts using PLMs~\cite{GaoFC20}, and mining prompts from a corpus~\cite{JiangXAN20}.

%However, these works on discovering discrete prompts are mainly focused on manually crafting the context to feed into PLMs, which is not only time consuming and non-intuitive, more importantly, models are highly sensitive to this context: improperly-constructed contexts cause artificially low performance~\cite{JiangXAN20}. Overcoming the need to manually specify prompts would make prompting a more widely useful analysis tool in the area of text generation.

\subsubsection{Continuous Prompts}

%Since the purpose of prompt-tuning is to help PLMs effectively perform downstream tasks, it is not necessary to limit the prompt to natural language text. 
Continuous prompts (\aka soft prompts), consisting of embedding vectors, are widely explored for text generation tasks. Two major advantages are expected: 1) relaxing the constraint that the prompt template should be natural language words; 2) removing the restriction that the template is parameterized by PLMs' parameters. Instead, continuous prompts have their own parameters that can be optimized based on training data of the text generation tasks. The most well-known method using continuous prompts for text generation is prefix-tuning~\cite{prefix-tuning}, which 
freezes the generative PLMs (\eg GPT-2, BART) and optimizes a sequence of task-specific vectors (called \textit{prefix}). In contrast to full-parameter fine-tuning, which requires storing a tuned copy of the model for each text generation task, prefix-tuning only optimizes the prefix for each text generation task. Similar to prefix-tuning, several studies used continuous prompts to solve other text generation tasks such as dialogue generation~\cite{abs-2111-02643}. 

\subsection{Property-Tuning for Text Generation}
\label{sec-property-tuning}
%For different text generation tasks, we need to maintain some specific properties 
%Different generation tasks usually need to maintain some specific properties in order to 

For different generation tasks, we need to consider specific language properties when tuning PLMs. In this section, we discuss three major properties that are widely desired for text generation.
%For example, in dialogue, we believe that the generated response should be relevant to the input dialog history and context. 

\subsubsection{Relevance}

According to the linguistic literature~\cite{li2021planning}, in text generation, \emph{relevance} means that the topical semantics conveyed in output text is highly related to the input text. As a representative example, in dialogue systems, the generated responses should be relevant to the historical utterances and other conditions, such as speaker persona and discourse topic. 

%which require the generated response to be relevant to the input history and other conditions such as the topic of dialog and persona of speakers. 

%In addition to the dialogue history, a condition corresponding to the type of response might be also provided as an external input such as the topic of dialogue and persona of speakers. The generated response should also be relevant to the conditions. 

%Recently, due to the absence of long-term memory and explicitly modeling the relation between input and output, RNN-based models still tend to generate irrelevant output text. (incorrect logic, rewriting) By contrast, PLMs pose a large number of parameters for storing long-term memory and utilize the powerful self-attention mechanism for connecting the input side and the output side. Therefore, applying PLMs to the dialog generation task would improve the relevance of generated text~\cite{transfertransfo,dialogpt}. 
Compared with traditional neural generative models, PLMs utilize more powerful multi-layer cross-attention mechanism to model the semantic associations between input and output, which can enhance the relevance of generated text to the input data (\eg the dialogue systems~\cite{transfertransfo,dialogpt}). 
A good example is  DialoGPT~\cite{dialogpt}  based on an auto-regressive language model GPT-2. Specially, DialoGPT was first trained on large-scale dialogue pairs/sessions, which could enable DialoGPT to capture the joint distribution of $\text{Pr}(history, response)$ in conversational flow for generating relevant responses to the history utterance.
%Specifically, TransferTransfo~\cite{transfertransfo} transfered PLMs to the task of open-domain dialogue systems, associated training and generation algorithms, which is able to significantly improve over the traditional Seq2Seq and information-retrieval models in terms of relevance of the response, coherence with the predefined personality and dialog history, and grammaticality and fluency as evaluated by automatic metrics. 
%Besides, DialoGPT~\cite{dialogpt} follow the OpenAI GPT-2 to model a multi- turn dialogue session as a long text and frame the generation task as language modeling, which was able to generate more relevant and context-consistent responses than traditional RNN-based models. 
Furthermore, Zeng~\etal~\cite{abs-2010-11140} utilized the masked language modeling objective to solve generate responses based on various types of dialogue context. Specifically, they proposed a TF-IDF based masking which selects more condition-related tokens to be masked, so that PLMs can generate condition-related expressions rather than the general language patterns. Besides, they adopted a non-parametric attention-based gating mechanism to switch between generating a general word or a condition-related word at each position.

\subsubsection{Faithfulness}

\emph{Faithfulness} is also an important language property to consider for text generation, which means the generated content should adhere to the semantics of input text. For example, text summarization aims to generate faithful text conveying the salient information of the input text. Faithfulness sometimes refers to the fact that the generated text is in accord with world facts. 

To generate faithful texts, PLMs should be able to accurately understand the core semantics of input and acquire sufficient world knowledge for solving the downstream task. 
It has been shown that PLMs have excellent natural language understanding capacities in capturing core semantics from plain text~\cite{bert}, and they indeed encode a large amount of world knowledge~\cite{JiangXAN20},  which is potentially beneficial to generate faithful summary by injecting background knowledge into text. 
%PLMs are able to understand natural language accurately which is helpful for PLMs to grasp the most critical information of text. Also, in pretraining, PLMs have encoded a lot of background knowledge into their massive amounts of parameters, which is potentially beneficial to generate faithful summary by injecting background knowledge into text. 
For example, Kryscinski~\etal~\cite{KryscinskiPXS18} utilized a contextual network in the PLM decoder to retrieve the most salient parts from the source document to improve the level of faithfulness of generated summaries. Besides, several studies proposed to generate faithful texts by introducing additional losses besides the text generation loss~\cite{RotheNS20,YangZGZ0D20}. Specifically, Yang~\etal~\cite{YangZGZ0D20} fine-tuned PLMs through a theme modeling loss which aims to make the generated summary semantically close to the original article for achieving faithful generation. %The denoising autoencoder can help PLMs extract salient information from corrupted text to further enhance the faithfulness of generated summaries.

\subsubsection{Order-Preservation}

In the NLP field, \emph{order-preservation} is a special property that refers that the order of semantic units (word, phrase, etc.) in both input and output text is consistent.
Such a property is key to several important text generation tasks, such as text paraphrasing and machine translation. In machine translation,  when translating from source language to target language, it often requires preserving some order of phrases in the source and target text for ensuring the accuracy of the translation results. 

%denotes the order of semantic units (word, phrase, etc.) in both input and output text is consistent. A representative example is machine translation. When translating from source language to target language, keeping the order of phrases consistent in the source and target text will ensure the accuracy of the translation results. 

In machine translation, word alignment is an extensively studied approach to achieve the order-preservation property.
%one line of research to improve the order-preservation property is to perform word alignment.
A representative study is Code-Switching Pre-training (CSP)~\cite{csp}. CSP first automatically extracted the word-pair alignment information from the source and target monolingual corpora. Then, to enhance the order-preservation property during translation, CSP continually pre-trained PLMs by predicting the sentence fragment on the source side given the aligned fragment in the target language. Moreover, to relax the restriction of discrete word alignment, another line of research aims to conduct continuous representation alignment to improve the order-preservation property. Wada~\etal~\cite{abs-1809-02306} focused on aligning word representations of each language by mapping word embeddings of each language into a common latent space. Lin~\etal~\cite{mrasp} proposed mRASP to enforce words and phrases that have similar meanings across multiple languages, to be aligned in the representation space.
%\xin{Although recent studies have achieved some progress on the English language, (what happend to English? and you seem to have mentioned multiple languages)} it is more challenging to enhance order-preservation across multiple languages. Thus, Lin~\etal~\cite{mrasp} proposed mRASP to enforce words and phrases that have similar meanings across multiple languages, to be aligned in the representation space.

%an approach to pretraining a universal multilingual machine translation model. The key to mRASP is the technique of randomly aligned substitution, which enforces words and phrases with similar meanings across multiple languages to be aligned in the representation space. Also, Wada~\etal~\cite{abs-1809-02306} focused on aligning word representations of each language by mapping word embeddings of each language into a common latent space, making it possible to preserve the word order consistent cross multiple languages.

\section{Challenges and Solutions}\label{sec:chal}

\begin{table*}[t]
	\centering
	\caption{Summary of major challenges in the three aspects and existing PLM-based solutions %in three different views. 
	}
	\label{tab:challenges}
	\small
	\begin{tabular}{c|c| p{0.68\textwidth}}
		\hline
		
		Aspect & Challenge & Solution \\
		
		\hline \hline
		
		\multirow{4}{*}{\tabincell{c}{Data\\Aspect}} & \multirow{2}{*}{\tabincell{c}{Lacking Enough\\Training Data}} & prior knowledge transfer~\cite{PengZLLLZG20,LiuZLRZY21,ZouZHG021}, data augmentation~\cite{XuWKL20,PasunuruCGXZBG21,abs-2109-08569,ChenY21a}, multi-task learning~\cite{GoodwinSD20,Bai0H20} \\
		\cline{2-3}
		
% 		& \multirow{2}{*}{\tabincell{c}{Domain\\Transfer}} & Continuously pretrained on specific out-domain data~\cite{ChenS21a,ZouZHG021}, or on auxiliary intermediate tasks~\cite{MauryaDKD21}.\\	
% 		\cline{2-3}
		
		& \multirow{2}{*}{\tabincell{c}{Bias in Pretraining\\Corpora}} & Mitigate the gender bias in word embeddings~\cite{BeutelCZC17}, identify and mask bias-sensitive tokens~\cite{DayanikP20}. \\
		
		\hline
		
		\multirow{4}{*}{\tabincell{c}{Model\\Aspect}} & \multirow{2}{*}{\tabincell{c}{Model\\Compression}} & Quantization by truncating PLMs weights~\cite{StockFGGGJJ21,ZadehEAM20}, pruning less critical weights~\cite{GordonDA20,abs-1909-12486,HouHSJCL20,FanGJ20}, knowledge distillation~\cite{ChenGCLL20,LiLZXYJ20,JiaoYSJCL0L20}. \\
		\cline{2-3}
		
		& \multirow{2}{*}{\tabincell{c}{Model\\Enhancement}} & Large-scale PLMs~\cite{gpt3,pangu,GShard,switch}, knowledge-enriched PLMs~\cite{causalbert,PetersNLSJSS19,ernie,HaoZP20}, efficient PLMs~\cite{HeLGC21,JiangYZCFY20}. \\
		\cline{2-3}
		
% 		& \multirow{2}{*}{\tabincell{c}{Model\\Robustness}} & Utilize character embeddings rather than sub-word embeddings~\cite{BoukkouriFLNZT20,MaCSLWH20}, adversarial data augmentation~\cite{JiaL17,WangB18,ZhouZWZWW21,XieDHL020}. \\
			
		\hline
		
		\multirow{4}{*}{\tabincell{c}{Optim.\\Aspect}} & \multirow{2}{*}{\tabincell{c}{Satisfying Text\\Properties}} & Enhance coherence~\cite{abs-1906-05743,li2021planning}, preserve factuality~\cite{ChenECLW20,LiTZWYW21,NanSZNMNZWAX20,DongWGCCL20}, improve controllable~\cite{pplm,KhalifaED21,PascualEMCW21}. \\
		\cline{2-3}
		
		& \multirow{2}{*}{\tabincell{c}{Mitigating Tuning\\Instabilities}} & Intermediate fine-tuning~\cite{abs-1811-01088,LiuWF21}, mixout strategy~\cite{LeeCK20}, supervised contrastive learning~\cite{GunelDCS21}. \\
		
		\hline
	\end{tabular}
\end{table*}

The three previous sections described three key aspects together with the basic methods used in PLM-based text generation. In this section, we further discuss the major challenges in each of the aspects and possible solutions. % about the above three aspects. 
%during the process of applying PLMs to text generation. In this section, we will present challenges and their solutions with respect to the three key points. 
A summary of these challenges and solutions is presented in Table~\ref{tab:challenges}. 

\subsection{Data Aspect}
We first discuss the challenges and solutions related to the data aspect. 

%In this part, we discuss the challenges and solutions related to data such as the pretraining corpus of PLMs and downstream training data for text generation.

\subsubsection{Lacking Sufficient Training Data}

In a number of text generation tasks, it is difficult to obtain sufficient annotated data. Transfer learning
provides an effective solution by transferring the knowledge of data-rich source tasks into data-scarce target text generation tasks. Besides, data augmentation and multi-task learning can be also used to address this problem.

%A commonly adopted approach is to plug the existing module with pretrained parameters. Then we fine-tune it with a few, one, or even no examples for the studied task, which are so-called few-shot, one-shot and zero-shot, respectively. 

\paratitle{Transfer Learning.}~To deal with the data scarcity issue, several studies proposed first fine-tuning PLMs on large amounts of external labeled corpora and then transferring into data-scarce target text generation tasks~\cite{PengZLLLZG20,LiuZLRZY21,ZouZHG021}. In particular, Peng~\etal~\cite{PengZLLLZG20} and Zou~\etal~\cite{ZouZHG021} first fine-tuned PLMs on substantial labeled dialog/summary data and then fine-tuned for the target dialog/summarization task in a new domain with limited labeled data. Similarly, Liu~\etal~\cite{LiuZLRZY21} first trained models on large-scale ungrounded dialogs and unstructured knowledge base separately to improve the low-resource knowledge-grounded dialog generation task. %Specifically, they devised a variant of Transformer with decoupled decoder which facilitates the disentangled learning of response generation and knowledge incorporation. 
%To deal with the low-resource dialogue summarization, Zou~\etal~\cite{ZouZHG021} proposed a multi-source pretraining paradigm to better leverage the external summary data. Specifically, we exploit large-scale in-domain non-summary data to separately pretrain the dialogue encoder and the summary decoder. %In multilingual translation, some low-resource languages lack sufficient parallel corpus. %Conneau~\etal~\cite{xlm} proposed XLM, which learns cross-lingual language models and can leverage the knowledge learned in high-resource languages to low-resource languages.

\paratitle{Data Augmentation.}~In recent literature, data augmentation has emerged as a critical method for increasing the amount of data by adding slightly modified copies of already existing data or newly created synthetic data from existing data. One line of research is to use retrieval models to obtain real data from external corpora as the augmented data~\cite{XuWKL20,PasunuruCGXZBG21}. %Xu~\etal~\cite{XuWKL20} combined a retrieval model with a language understanding model to create data from open-domain texts, while 
For the query-focused summarization task, Pasunuru~\etal~\cite{PasunuruCGXZBG21} used a search engine, \ie Bing, to retrieve the answer paragraph as the synthetic summary and used the top ranked documents as input text. Another line of work is to use perturbation-based methods by corrupting the original text~\cite{abs-2109-08569,ChenY21a}. For example, Chen~\etal~\cite{ChenY21a} presented a set of data augmentation methods for conversation summarization, such as random swapping/deletion to randomly swap or delete utterances in conversations. %, dialogue-acts-guided insertion to interrupt the development of conversations, and conditional-generation-based substitution to substitute utterances with their paraphrases.

\paratitle{Multi-Task Learning.}~Leveraging other data-rich tasks and datasets can also overcome the data scarcity issue.
%Because of the difficulty in obtaining sufficient data in the target task, researchers consider exploring multi-task learning by 
%leveraging other data-rich tasks and datasets. 
Most studies usually incorporated similar auxiliary generation tasks for enhancing the primary text generation task~\cite{GoodwinSD20}. However, these methods usually adopt independent decoders for each task, thus breaking the semantic connections between high- and low-resource text generation tasks. To bridge this gap, Bai~\etal~\cite{Bai0H20} employed a unified decoder which learns the alignments and patterns across multiple languages in machine translation. 

\subsubsection{Data Bias from Pre-training Corpora}
%Since PLMs are pretrained on the unlabeled corpus collected from the Web, it might incorporate the data bias of the pretraining corpus. 

%It has been found that the the generated text from these PLMs are likely to be biased towards some attributes~\cite{}, \ie may favor a particular race, gender or aged people, which is not desired for the target generation tasks. 

In sociology, bias is an unjustified prejudice in favour of or against a person, group, or thing~\cite{garrido2021survey}. PLMs are generally trained using real-world data in such a way that they model the statistical properties of the training data. As a result, they inherit the biases and stereotypes that are common in the data~\cite{garrido2021survey}. These biases and stereotypes can pose significant challenges in downstream text generation tasks~\cite{gpt3}.

%such specific data possibly show bias and PLMs are prone to learn and amplify the bias. As a result, the generated text from these PLMs are biased \ie may favor a particular race, gender, or aged people. 
It has been shown that the generated texts from PLMs are likely to be biased towards some attributes~\cite{gpt3}, \ie  favoring a particular race, gender or aged people, which is not desired for the text generation tasks. These undesirable biases are unexpectedly hidden in model components such as word embeddings~\cite{BolukbasiCZSK16} and attention heads~\cite{VigGBQNSS20}.
%Several studies have shown that the pretraining corpus has lead to undesirable bias (\eg gender bias) in model components such as word embedding~\cite{BolukbasiCZSK16} and attention heads~\cite{VigGBQNSS20}.
%More recent, state-of-the-art word embeddings (\eg ELMo, BERT, and GPT-2) are generally contextual, where the vector representation of a word from the trained model is dependent on the context around the word. However, it has been found that word embeddings could present undesirable bias. Bolukbasi~\etal~\cite{BolukbasiCZSK16} present evidence of gender bias in word2vec embeddings, along with proposing a method for removing bias from gender-neutral terms. Vig~\etal~\cite{VigGBQNSS20} investigate which model components (attention heads) are responsible for gender bias in transformer-based language models (GPT-2). 
A simple approach to mitigating the gender bias in word embeddings is to ``swap'' gendered terms in training data when generating word embeddings~\cite{ZhaoZLWC18}. %Beutel~\etal~\cite{BeutelCZC17} develop an adversarial system for debiasing language models by relating the distribution of training data to its effects on properties of fairness in the adversarial system. 
Furthermore, simply masking names and pronouns may also reduce biases and improve the performance of certain language tasks~\cite{DayanikP20}. 
%Minot~\etal~\cite{abs-2103-05841} proposed a data augmentation-based approach to reduce gender bias while Liang~\etal~\cite{LiangWMS21} proposed A-INLP approach which dynamically identifies bias-sensitive tokens. 
However, to date, there is still no general, unified approach to reducing the data bias from PLMs for text generation. 
Some of these techniques for bias detection and mitigation have been critiqued as merely capturing over-simplified dimensions of bias with proper debiasing requiring more holistic evaluation~\cite{GonenG19}. %According to Liang~\etal~\cite{LiangWMS21}, existing approaches towards mitigating biases in text generation require retraining the models through adversarial trigger prompts~\cite{ShengCNP20}, data augmentation or collection~\cite{DinanFWUKW20}, and different objective functions~\cite{QianMZH19}. %These various approaches have also been applied to image captioning~\cite{HendricksBSDR18}, image retrieval~\cite{OtterbacherCDC18}, and dialog~\cite{LiuDFLLT20}. 
%However, these debiasing approaches cannot applied to large PLMs since it is difficult to retrain a new LM whenever a new source of bias is uncovered from data~\cite{LiangWMS21}.

\subsection{Model Aspect}

In this section, we present the challenges from the architecture design, and discuss corresponding solutions for text generation.

\subsubsection{Model Compression}

Although PLMs have achieved great success on text generation, the backbone Transformers are still bulky and resource-hungry, resulting in high memory consumption, computational overhead, and energy cost. To address these issues, more and more approaches are proposed to compress PLMs~\cite{compression_survey}, such as quantization, pruning, and knowledge distillation. %These compression methods can adapted to all types of PLMs. 

\paratitle{Quantization.} Quantization means reducing the number of unique values used to represent PLMs weights, which in turn allows to represent them using fewer bits~\cite{compression_survey}. As most PLMs are built upon Transformer, quantization can be generally applied to those weights residing in fully-connected layers (\ie embedding layers, linear layers, and feed-forward network layers). 
%Simply truncating PLMs weights to the target bitwidth may incur a sizable drop in accuracy since truncation makes certain weights go through a severe drift in their value~\cite{StockFGGGJJ21}.
%A naive approach is to truncate PLMs weights to the target bitwidth. However, this may incur a sizable drop in accuracy since truncation makes certain weights go through a severe drift in their value, known as quantization noise~\cite{StockFGGGJJ21}.
However, when the model parameters are compressed, the generation capacity might be reduced. 
To alleviate the issue of generating unsatisfactory text with truncated PLMs, a promising solution is to first identify important weights and then avoid truncating them during the quantization step~\cite{ZadehEAM20}. %For example, Zadeh~\etal~\cite{ZadehEAM20} assume the PLMs weight matrix follow Gaussian distribution and then identify outliers. Then, by not quantizing these outliers, they are able to perform post-training quantization without any retraining requirements. %Moreover, Quantization-Aware Training (QAT) proposes to adjust quantized PLMs weights by involving additional training steps, including fixed-length integer quantization~\cite{ZafrirBIW19,BooS20}, Hessian-based mixed-precision quantization~\cite{ShenDYMYGMK20}, and noise-based quantization~\cite{StockFGGGJJ21}.

\paratitle{Pruning.} Pruning refers to identifying and removing redundant and/or less important weights~\cite{compression_survey}. %This compression method sometimes can make PLMs more robust. 
Pruning methods for text generation largely fall into two categories~\cite{compression_survey}. The first type of unstructured pruning prunes individual weights by locating the set of least important weights in PLMs. The importance of weights can be measured by specific metrics such as absolute values~\cite{GordonDA20} and gradients~\cite{abs-1909-12486}. %According to Ganesh \etal~\cite{compression_survey}, these pruning methods include magnitude weight pruning~\cite{GordonDA20,MaoWWZWZYTB20}, which just removes weights close to zero, movement-based pruning~\cite{Sanh0R20,TambeHPJYDSWR0W21}, which removes weights shifting towards zero during fine-tuning, and reweighted proximal pruning~\cite{abs-1909-12486}, which applies an iteratively reweighted $\ell_1$ minimization for decoupling pruning and error back-propagation. 
%Since unstructured pruning considers each PLMs weight individually, the set of pruned weights can be arbitrary and irregular. Although it might decline the model size, the improvement in run-time memory or speed is negligible. 
The second type of structured pruning prunes structured blocks of weights or even complete components of PLMs by reducing and simplifying certain  modules such as attention heads~\cite{HouHSJCL20} and Transformer layers~\cite{FanGJ20}. %It has been reported that only 1-2 attention heads per encoder unit may also achieve high accuracy, even though the original PLMs usually has 16 attention heads~\cite{MichelLN19}. Therefore, Hou~\etal~\cite{HouHSJCL20} propose to randomly prune attention heads during the training phase, which can create a model robust to various numbers of attention heads. Besides, Fan~\etal~\cite{FanGJ20} also propose to drop Transformer layers randomly during inference.

\paratitle{Knowledge Distillation.} Knowledge distillation refers to training a smaller model (called the \textit{student}) using the output of PLMs (called the \textit{teacher}). %The student model can learn from multiple kinds of outputs of the teacher model through specific loss functions such as cross-entropy, KL divergence, and MAE.
%such as the logits in the final layer of PLMs, the outputs of the encoder units, and the attention maps. Moreover, there are multiple forms of loss functions for knowledge distillation such as cross-entropy loss, KL divergence, and MAE. 
First, the student model can directly learn from the output word distribution of the final softmax layer in PLMs, which allows the student to mimic the generated text of the teacher by replicating the word distribution across the whole vocabulary~\cite{ChenGCLL20}. 
%The student can apply a totally different architecture from the teacher (PLMs) while does not need to  adopt a smaller Transformer. 
Second, the student can also learn from the output tensors of PLMs encoders~\cite{LiLZXYJ20}. Intuitively, the representations of PLMs encoder may contain meaningful semantics and contextual relationships between input tokens, which is helpful for generating accurate text. Third, %attention map is the softmax distribution output of the self-attention layers, which indicates the contextual dependence between the input tokens. 
%In literatures~\cite{ClarkKLM19}, it has been reported that attention maps in PLMs can identify distinguishable linguistic relations, \eg verbs and corresponding objects, pronouns and corresponding nouns. 
by replicating attention distributions between input data and output text, the student can also learn the contextual dependency between input and output~\cite{JiaoYSJCL0L20}. %A common distillation method learning from attention maps is to directly minimize the difference between the student and the teacher multi-head self-attention outputs. %Previous work also proposed to replicating only the last attention map in the teacher in order to fully capture the contextual dependece~\cite{WangW0B0020}.

\subsubsection{Model Enhancement}

Although PLMs have achieved great success nowadays, they are still far from our expectations. Recently, there has been a surge of interest in the research community to strengthen existing PLMs to improve the performance of text generation.

\paratitle{Large-scale PLMs.} %In general, the sizes of PLMs are in the range of 110M to 340M~\cite{bert,gpt2,bart}. Since the advent of GPT-3~\cite{gpt3}, the parameters of PLMs become billions of scale. 
Kaplan~\etal~\cite{abs-2001-08361} have shown that the performance of PLMs can be boosted by scaling up the amount of PLMs' parameters. This observation sparked the development of large-scale PLMs in text generation~\cite{gpt3,pangu}.
%According to Kaplan~\etal~\cite{abs-2001-08361}, the performance of PLMs can be improved by increasing the size of the model or training the model on much large amount of corpus or training the model for more training steps. 
The most representative large-scale PLMs for text generation is GPT-3~\cite{gpt3}, which contains 175 billion parameters, 10x more than any previous non-sparse PLMs. With a large number of parameters, GPT-3 can achieve strong performance in various text generation tasks without any gradient updates or fine-tuning.

\paratitle{Knowledge-Enriched PLMs.} %During pretraining, PLMs learns the knowledge available in the large-scale pretraining corpus by utilizing several pretraining tasks. 
Recent research has found that integrating knowledge from external knowledge sources can enhance the text generation performance of PLMs~\cite{ernie3,calm}. %Some of the popular external knowledge sources are WordNet, Wikidata in the general domain and UMLS~\cite{Bodenreider04} in specific domains like Biomedical. 
Specifically, ERNIE 3.0~\cite{ernie3} was pretrained on a 4TB corpus consisting of plain texts and a large-scale knowledge graph for both language understanding and generation tasks. Without incorporating explicit knowledge, CALM~\cite{calm} can  encode commonsense knowledge into parameters by teaching PLMs to write and reason with common concepts through pre-training strategies, yielding better performance on text generation tasks. 
%Some of the examples of knowledge enriched PLMs are CausalBERT~\cite{causalbert}, KnowBERT~\cite{PetersNLSJSS19}, %SenseBERT~\cite{LevineLDRPSSSS20}, 
%ERNIE~\cite{ernie} %, E-BERT~\cite{PornerWS20} 
%in the general domain and clinical Kb-BERT~\cite{HaoZP20} %CoderBERT~\cite{coderbert} 
%in specific domain like Biomedical. Specifically, KnowBERT~\cite{PetersNLSJSS19} integrates WordNet and a subset of Wikipedia into BERT, and ERNIE~\cite{ernie} utilize both large-scale textual corpora and KGs.  %LiBERT~\cite{LauscherVPKG20} is pretrained from scratch using a novel lingual relation classification (LRC) task along with MLM and NSP. LRC task can allow PLMs to integrate linguistic knowledge. 
%While for text generation, there is still no knowledge-enriched PLMs. We guess the reason might be that the text generation task requires the model to consider a wide range of knowledge to generate diverse text.

\paratitle{Efficient PLMs.} Pre-training PLMs on large-scale text data is prohibitively expensive. Recently, it has been demonstrated that by meticulously structuring the model architecture, it is possible to obtain equivalent or higher text generation performance with less pre-training data~\cite{calm} or lower pre-training costs~\cite{JiangYZCFY20}. For example,
CALM~\cite{calm} developed a mutually reinforced pre-training framework with generative and contrastive objectives, thus achieving comparable results to other larger PLMs such as T5 while only being pre-trained on a small corpus for a few steps.

\subsection{Optimization Aspect}

In this part, we discuss challenges and solutions about the optimization of PLMs for text generation.

\subsubsection{Satisfying Special Text Properties}
In Section~\ref{sec-property-tuning}, we introduced three basic text properties. In this section, we will present three more difficult properties for text generation tasks, \ie coherence, factuality, and controllability.
%It is meaningful to consider some special concerns such as language coherence and text fidelity in specific generation tasks when fine-tuning PLMs.

\paratitle{Coherence.}~In linguistics~\cite{LiH14a}, language coherence is what makes a multi-sentence text meaningful, both logically and syntactically.
An essential technique to improving coherence is to elaborately plan the generated content, which is known as text planning~\cite{li2021planning,HuaS020}. For example, Li~\etal~\cite{li2021planning} designed a text generation model based on a two-level text plan: (1) the document plan is modeled as a sequence of
sentence plans in order, and (2) the sentence plan is modeled as
an entity-based subgraph from KG. The local coherence is naturally enforced by KG subgraphs, and the global coherence can be improved by generating a coherent sequence of subgraphs. %During the generation process, decoding strategies such as top-$k$ sampling usually produce improper tokens especially at the boundary of sentences, reducing the discourse coherence. 
Wang~\etal~\cite{WangLZ21} proposed a two-stage planning, \ie the first stage is to organize the story outline which illustrates the story plots and events, and the second stage is to expand the outline into a complete story.

\paratitle{Factuality.} The input data (\eg infobox) for text generation tasks (\eg table-to-text generation) usually contains some factual information. In such cases, the generated content should adhere to the original input facts. However, lacking direct access to the input facts or explicit supervision makes PLMs unable to retain text factuality in generation process. For data-to-text generation, the pointer generator~\cite{pointer} is usually adopted to copy the input facts into output for preserving factuality~\cite{ChenECLW20,LiTZWYW21}. %For example, Li~\etal~\cite{LiTZWYW21} adopted pointer-generator to copy entities from input knowledge data.
%Gong~\etal~\cite{tablegpt} proposed to utilize multi-task learning, in order to reconstruct from table embeddings and enforce the match between table embeddings and content embeddings. Besides, the pointer generator~\cite{pointer} can be applied to KG-to-text generation to copy the entity and relation information in KG~\cite{ChenECLW20}. Cao~\etal~\cite{Cao021} presented a contrastive learning framework by leveraging both reference summaries, as positive training data, and automatically generated erroneous summaries, as negative training data, to train summarization systems that are better at distinguishing between them. 
Furthermore, to make summarization models produce more factual summaries, some studies proposed  evaluation metrics or correction methods to measure and revise the generated text for preserving factuality~\cite{NanSZNMNZWAX20,DongWGCCL20}. %Specifically, Nan~\etal~\cite{NanSZNMNZWAX20} proposed an automatic evaluation metric to measure factual consistency, and then used a learning algorithm that maximizes the metric during model training. Dong~\etal~\cite{DongWGCCL20} proposed a suite of two factual correction models that leverages knowledge learned from question answering models to make corrections in system-generated summaries via span selection. %To better understand the potential threats of synthetic news, Shu~\etal~\cite{ShuLDL21} retrieves external facts to enrich the output and reconstructs the input claim from the generated content to improve the consistency among the input and the output. 

\paratitle{Controllability.}  
In text generation, many applications need a good control over the output text. For example, to generate reading materials for kids, we would like to guide the output stories to be safe, educational and easily understandable by children. %Amplayo~\etal~\cite{AmplayoAL21} construct a synthetic training dataset consisting of reviews, a pseudo-summary, and three types of aspect controllers which reflect different levels of granularity: aspect-related keywords, review sentences, and document-level aspect codes to generate aspect-controllable summaries. 
The Plug and Play Language Model, also known as PPLM~\cite{pplm}, is an example of a controllable PLM that combines a PLM with one or more simple attribute classifiers that direct text generation without further PLM training. %Similarly, Yang~\etal~\cite{Yang0XLBWWL20} leveraged a style classifier to fine-tune PLMs in order to steer response generation towards the target style in both a word-level and a sentence-level.
%More flexible, via the sequence span rewriting objective, SSR~\cite{ssr} can specify which part of text to be rewritten for controllable text generation.
Several studies achieved controllablility from a distributional view~\cite{KhalifaED21,PascualEMCW21}. %Khalifa~\etal~\cite{KhalifaED21} formalized controllable generation as a constraint satisfaction problem combined with a divergence minimization objective, which provides a single framework both for ``distributional'' constraints (collective statistical requirements) and for ``pointwise'' constraints (hard requirements on each individual). 
Pascual~\etal~\cite{PascualEMCW21} described a plug-and-play decoding approach in a single sentence: given a topic or keyword, the model adds a shift to the probability distribution over the vocabulary towards semantically similar words. %Singh~\etal~\cite{GoswamySBM20} posit a model built on top of GPT-2 capable of generating affect-driven and topic focused sentences without losing grammatical correctness as the affect intensity increases. Their model provides degrees of freedom in terms of the choice of the base text generation model, the emotion category, with fine-grained control over emotion intensity for each category, and the topic of the generated text. 
%Yu~\etal~\cite{YuYS21} propose Attribute Alignment to infuse attribute representations into a unconditional PLM without changing the PLM parameters. Since it does not involve training LM parameters, they can do controlled text generation without sacrificing the linguistic quality of the original LM. 
%Chang~\etal~\cite{ChangYIM21} design a framework that displays multiple candidate upcoming topics, of which a user can select a subset to guide the generation. This framework consists of two components: (1) a method that produces a set of candidate topics by predicting the centers of word clusters in the possible continuations, and (2) a text generation model whose output adheres to the chosen topics. %Liu~\etal~\cite{LiuSLSBSC20} combine an out-of-the-box pretrained (``base'') LM with ``expert'' LMs and/or ``anti-expert'' LMs, which model text with desirable and undesirable attributes, respectively. By generatively modeling text with particular attributes and directly combining the output distributions from each LM, their model leverages subtle signals expressible by language models for effective attribute control. %Dathathri~\etal~\cite{pplm} propose a simple alternative: the Plug and Play Language Model (PPLM) for controllable language generation, which combines a pretrained LM with one or more simple attribute classifiers that guide text generation without any further training of the LM.

\subsubsection{Mitigating Tuning Instabilities}

Due to the catastrophic forgetting nature of PLMs and small size of text generation datasets, tuning PLMs for text generation is usually unstable \ie fine-tuning the model with different random seeds results in a wide variance of performance. %Mosbach~\etal~\cite{MosbachAK21} demonstrated that tuning instability can be attributed to a) optimization difficulties which lead to vanishing gradients and b) generalization issues. 
The possible solutions include intermediate fine-tuning, mixout and using supervised contrastive loss.

\paratitle{Intermediate Fine-Tuning.} Recent studies have shown that first training PLMs on data-rich intermediate labeled datasets (\eg a similar NLP task from the same target domain) before fine-tuning them on data-scarce target text generation tasks can achieve better performance in target tasks~\cite{abs-1811-01088,LiuWF21}. %Intuitively, large-scale unsupervised pretraining could impart PLMs with substantial knowledge of the target tasks. 
%Phang~\etal~\cite{abs-1811-01088} explored the usage of a second stage of pretraining with data-rich intermediate supervised tasks, improving both the robustness and effectiveness of the resulting target task model. %To address the problem of unseen languages in the translation pairs, 
For example, Liu~\etal~\cite{LiuWF21} constructed an intermediate monolingual corpus of the target language (\eg Kazakh) and fine-tuned mBART to reconstruct the corrupted monolingual text for improving the translation quality of the low-resource target language.
%to cover both the source and target languages in the subsequent fine-tuning phase. Here the intermediate dataset (monolingual corpus of the target language) and target datasets (target language pairs) are from the same domain \ie low-resource target language but from different tasks. Based on the same domain dataset, Xing~\etal~\cite{XingW21} introduced three self-supervised tasks to help PLMs understand the structured input table. The first two tasks are only used to train the encoder while keeping the decoder frozen, and the third task are used to train the encoder and decoder at the same time.

\paratitle{Mixout Strategy.} When fine-tuning PLMs, dropout~\cite{dropout} has been used as a regularization method to prevent performance degeneration if there are only a small number of training instances. Lee~\etal~\cite{LeeCK20} introduced a variant of dropout, mixout, which stochastically mixes parameters of two PLMs. %The mixout technique is motivated by dropout. 
The mixout strategy can regularize learning by minimizing the deviation from one of the two PLMs and the strength of regularization adapts along the optimization trajectory.

\paratitle{Contrastive Learning.} The most used cross-entropy loss in text generation, \ie the KL-divergence between one-hot vectors of labels and the distribution of model’s outputs, lacks robustness to noise labels~\cite{ZhangS18} or adversarial examples~\cite{ElsayedKMRB18}. Thus, fine-tuning PLMs with cross-entropy loss tends to be unstable, especially when labeled data is limited. An effective solution is to capture the similarity between examples in one class and contrast them with examples in other classes~\cite{GunelDCS21}. To this end, Gunel~\etal~\cite{GunelDCS21} combined the cross-entropy loss with a supervised contrastive learning loss that pushes the words from the same class close and the words from different classes further apart.

\section{Evaluation and Resources}\label{sec:eva}

In this section, we will discuss several commonly used evaluation metrics and resources with respect to PLMs for text generation.

\subsection{Evaluation}

With the growing variety of text generation applications and datasets, there are several advantages of automatic evaluation: it is potentially much cheaper and quicker than human evaluation, and it is repeatable~\cite{BelzR06}. Therefore, we mainly concentrate on automatic evaluation metrics for text generation in this part. Following Celikyilmaz~\etal~\cite{eva_survey}, we present four categories of metrics, \ie $n$-gram overlap metrics, diversity metrics, semantic similarity metrics, and logit-based metrics. We list the metrics used in each text generation task in Table~\ref{tab:eval}.

\subsubsection{N-Gram Overlap Metrics}

These metrics measure the degree of word ``matching'' between machine-generated and ground-truth texts at the word level. 

\paratitle{BLEU.} The Bilingual Evaluation Understudy (BLEU)~\cite{bleu} is one of the first metrics used to compare the similarity of two sentences. This metric was originally proposed for machine translation by comparing a candidate translation of text with one or more reference translations and now applied in various generation tasks. BLEU-$n$ measures the precision of the co-occurrences of $n$-grams between the generated and real text and conducts length penalty on shorter generated text. Specially, SacreBLEU~\cite{sacrebleu} is recommended for use in machine translation to avoid inconsistency issue. Several smoothing methods~\cite{smoothingbleu} are also proposed to evaluate short sentences.

\paratitle{ROUGE.} Recall-Oriented Understudy for Gisting Evaluation (ROUGE)~\cite{rouge} is a set of metrics for measuring automatic summarization of long texts consisting of multiple sentences. ROUGE-$n$ counts the F1 score of the overlapping $n$-grams between generated and ground-truth texts. 

\paratitle{METEOR.} The Metric for Evaluation of Translation with Explicit ORdering (METEOR)~\cite{meteor} is proposed to address some issues found in BLEU. Compared to BLEU, METEOR is computed based on the harmonic mean of the unigram precision and recall, and measures word-to-word matches between generated and real text based on WordNet.

\paratitle{ChrF++.}  Character $n$-gram F-score (ChrF++)~\cite{chrf++} is an automatic evaluation metric for machine translation. Different from the word level co-occurrence of BLEU, ChrF++ is mainly focused on the character-level matching so as to consider morpheme overlapping.

% \paratitle{CIDEr~\cite{VedantamZP15}.} Consensus-based Image Description Evaluation (CIDEr) is an automatic metric for measuring the similarity of a generated sentence against a set of human-written sentences using a consensus-based protocol. The metric is firstly proposed for image caption, showing high agreement with consensus as assessed by humans.

\subsubsection{Diversity Metrics}

Lexical diversity is desirable in many text generation tasks, such as dialogue systems and story generation. For these tasks, it is necessary to conduct diversity evaluation on generated texts.

\paratitle{Distinct.} Distinct-$n$ measures the degree of diversity by calculating the number of distinct $n$-grams in generated text~\cite{LiGBGD16}. This metric is scaled by total number of generated tokens to avoid favoring long sentences.

% \paratitle{Self-BLEU.} In a generated text collection, regarding one generated text as hypothesis and the other generated texts as reference, we can calculate BLEU score for every generated text, and define the average score to be the Self-BLEU score~\cite{ZhuLZGZWY18}. A higher Self-BLEU score implies less diversity of the generated results from a model.

\subsubsection{Semantic Similarity Metrics} 
The above metrics are focused on the literal word comparison. Many studies also proposed to compare the implicit semantics between generated text and ground-truth text. A typical approach is to map both generated text and ground-truth text into sentence vectors and then compare their embedding similarity. 
%The researchers used neural networks to capture semantic meaning and syntactic structure of sentences by mapping them into vectors, and the results of text generation can be evaluated using sentence embeddings from the generated and reference texts. 
%Semantic similarity also include tf-idf cosine similarity. You need to explain both your setting and global meaning of semantic similarity.

\paratitle{BERTScore.} Given the excellent performance of BERT across many tasks, BERTScore~\cite{ZhangKWWA20} leverages the pre-trained contextual embeddings from BERT and compares words in candidate and reference texts by cosine similarity. BERTScore has  proven to correspond well with human judgments on sentence-level and system-level evaluations~\cite{eva_survey}.

\subsubsection{\textbf{Logit-Based Metrics}} 

In text generation, the probability of a generated text $y=\langle y_1,\dots,y_n \rangle$ can be formulated as $\text{Pr}(y) = \prod_{j=1}^{n}\text{Pr}(y_j|y_{1:j-1};x)$, where $x$ denotes the input data, and $y_{1:j-1}$ denotes the previous tokens $\langle y_1,\dots,y_{j-1} \rangle$. 
Logit-based metrics evaluate the generated text from a probabilistic view.

% \paratitle{NLL.} Negative log-likelihood (NLL) is originally introduced in SeqGAN~\cite{YuZWY17} to tell how good the generated data is fitted by the oracle language model. In SeqGAN, a randomly initialized LSTM is regarded as a true model, and the text generation model needs to minimize the average negative log-likelihood of generate data on oracle LSTM, \ie $\mathbb{E}_{y \sim q}\log \text{Pr}(y)$, where $y$ denotes the generated text. Since an LSTM is regarded as a true model, NLL can calculate the average loss on every sentence, word by word:
% \begin{equation}
% 	\text{NLL}(y) = -\mathbb{E}_{y \sim G_\theta} \sum_{j=1}^n \log(G_{oracle}(y_j|y_{1:j-1})),
% \end{equation}
% where $G_{oracle}$ denotes the oracle LSTM, and $G_\theta$ denotes the generative model. 

\paratitle{PPL.} In information theory, perplexity (PPL) is a measurement of how well a probability distribution or probability model predicts a sample compared with the ground-truth~\cite{BrownPPLM92}. A low perplexity indicates the probability distribution is good at predicting the sample. Therefore, the perplexity of the discrete probability distribution $\text{Pr}(\cdot)$ is defined as:

\begin{equation}
	\text{PPL}(\text{Pr}(y)) \coloneqq e^{\text{H}(\text{Pr}(y))} = e^{-\sum_y \text{Pr}(y)\ln \text{Pr}(y)} = \prod_y \text{Pr}(y)^{-\text{Pr}(y)},
\end{equation}
where $\text{H}(\text{Pr}(y))$ is the entropy of the distribution $\text{Pr}(\cdot)$.

\subsection{Resources}

In this section, we will introduce some available open-source libraries and benchmarks.

\subsubsection{Open-Source Libraries}

There are a number of public text generation libraries that can be used to implement PLM-based text generation models. Transformers~\cite{WolfDSCDMCRLFDS20} is an all-featured library for Transformer-based PLMs, and Fairseq~\cite{OttEBFGNGA19} is a library to train custom models for translation, summarization, language modeling and other text generation tasks. Besides, some of libraries like FastSeq~\cite{fastseq}, DeepSpeed~\cite{RasleyRRH20}, and LightSeq~\cite{WangXWWL21} are useful to increase the inference speed of models. TextBox~\cite{li2021textbox} supports 21 text generation models, including several prevalent PLMs, and diverse generation strategies (\eg top-$k$, beam search) and evaluation metrics (\eg BLEU, Distinct). One can %Users are convenient to 
easily choose different PLMs, optimization methods, and evaluation metrics by setting corresponding hyper-parameters with just a few lines of code.  

%Some of libraries like TextBox~\cite{li2021textbox} and Trankit~\cite{nguyen2021trankit}
%which are built on the top of Transformers library make constructing text generation models easier with just a few lines of code. 

%\xin{I suggest you add more words about TextBox, better align the above three aspects with the implementations in TextBox. E.g., how to set the model achitecture and how to optimize the PLMs. }

\subsubsection{Evaluation Benchmarks}
In order to evaluate the comprehensive capacities of PLMs, several important evaluation benchmarks are created and released, which involve multiple evaluation tasks from different aspects. 
%PLMs have made great progress in a host of Natural Language Understanding (NLU) tasks. Meanwhile, the development of general evaluation benchmarks has also helped drive the progress of these PLMs. 
In addition to GLUE~\cite{glue} and SuperGLUE~\cite{superglue} which are general language understanding evaluation benchmarks, an increasing number of general benchmarks targeted for text generation have recently been proposed. Liu~\etal~\cite{glge} introduced the General Language Generation Evaluation (GLGE) benchmark, a new multi-task benchmark for evaluating the generalization capabilities of text generation. GLGE contains 8 English language generation tasks, covering summarization, question generation, generative question answering, and dialogue. For each task, GLGE designs three sub-tasks in terms of task difficulty (\ie GLGE-Easy, GLGE-Medium, and GLGE-Hard).

\section{Application}\label{sec:app}

% \usepackage{multirow}
\begin{savenotes}
\begin{table}[t]
\small
\centering
\caption{A summary of common datasets and metrics used in each generation task. \textsuperscript{\dag}BLEU with smoothing method 7 (with NLTK version 3.4) is usually employed in open-domain dialogue system~\cite{plato}. \textsuperscript{\ddag}Inform (rate) and Success (rate) are two accuracy metrics specially designed for task-oriented dialogue system~\cite{BudzianowskiV19}.}
\label{tab:eval}
\begin{tabular}{c|c|c|c}
\hline
Tasks & Sub-Tasks & Datasets & Metrics \\ \hline \hline
\multirow{2}{*}{\tabincell{c}{Machine\\ Translation}} & \tabincell{c}{Unsupervised MT} & \multirow{2}{*}{\tabincell{l}{WMT'14 English-French~\cite{xlm},\\ WMT'16 German-English~\cite{xlm}}} & \multirow{2}{*}{SacreBLEU} \\ \cline{2-2}
 & \tabincell{c}{Supervised MT} &  &  \\ \hline
\multirow{3}{*}{Summarization} & \tabincell{c}{Vanilla\\ Summarization} & \tabincell{l}{CNN/DailyMail~\cite{mass}, \\ XSum~\cite{mass}, GigaWord~\cite{mass}} & ROUGE, BERTScore \\ \cline{2-4} 
% & Long Doc. Sum. & arXiv~\cite{GidiotisT20}, PubMed~\cite{GidiotisT20} & ROUGE \\ \cline{2-4} 
 & \tabincell{c}{Dialogue\\ Summarization} & SAMSum~\cite{ChenY20} & ROUGE \\ \cline{2-4} 
% & Multilingual Sum. & GigaWord~\cite{xnlg}, MLGSum~\cite{marge} & ROUGE \\ 
\hline
\multirow{3}{*}{\tabincell{c}{Dialogue\\ System}} & \tabincell{c}{Open-Domain\\ Dialogue System} & \tabincell{l}{PersonaChat~\cite{plato}, DailyDialogue~\cite{plato},\\ DSTC7-AVSD~\cite{plato}} & \tabincell{c}{Perplexity\\ BLEU\textsuperscript{\dag}, Distinct} \\ \cline{2-4} 
 & \tabincell{c}{Task-Oriented\\ Dialogue System} & MultiWOZ~\cite{BudzianowskiV19} & BLEU, Inform\textsuperscript{\ddag}, Success\textsuperscript{\ddag} \\ \hline
\multicolumn{2}{c|}{Question Generation} & SQuAD~\cite{unilm} & BLEU, ROUGE, METEOR \\ \hline
\multicolumn{2}{c|}{Story Generation} & \tabincell{l}{ROCStories~\cite{GuanHHZZ20},\\ WritingPrompts~\cite{plotmachines}} & \tabincell{c}{Perplexity\\ BLEU, Distinct} \\ \hline
\multicolumn{2}{c|}{Data-to-text Generation} & \tabincell{l}{AGENDA~\cite{Ribeiro2020}, LDC2017T10~\cite{MagerANSLFR20},\\ WikiBio~\cite{ChenECLW20}, WebNLG~\cite{Ribeiro2020}, E2E~\cite{kgpt}} & \tabincell{c}{BLEU, ROUGE,\\ METEOR, chrF++} \\ \hline
\end{tabular}
\end{table}
\end{savenotes}

As discussed in Section~\ref{sec:back}, text generation can be instantiated into different kinds of applications. To summarize existing text generation applications, we present an overview of different tasks (as well as corresponding common datasets and metrics) in Table~\ref{tab:eval}.
%And PLMs have been successfully applied to these text generation applications. The overall taxonomy of these tasks is shown in Table~\ref{tab:eval}, as well as corresponding common datasets and metrics.
In what follows, we will highlight three classic applications, \ie machine translation, text summarization and dialogue system, and briefly discuss how to design a task-specific PLM to adapt to specific text generation tasks.

\subsection{Machine Translation}
Machine translation (MT) is the process of automatically translating one language into another. With the advent of deep learning, Neural Machine Translation (NMT) has emerged as the dominant method in both academic research and commercial use~\cite{mnmt_survey}. 
Machine translation can be classified into two types: unsupervised machine translation and supervised machine translation, depending on whether parallel corpora are available  for fine-tuning PLMs.

\subsubsection{Unsupervised Machine Translation}
Unsupervised Machine Translation (UMT) refers to the use of solely monolingual corpora without any parallel data for both pre-training and fine-tuning PLMs. UMT enables machine translation to no longer rely on large-scale annotated corpora, and also brings remarkable advances in low-resource language translation. When using PLMs for UMT, there are typically two steps involved~\cite{umt}: 1) PLMs are pre-trained on monolingual corpora in a variety of languages, learning word embeddings and modeling probabilities for each sentence in each language; 2) Iterative back-translation is then leveraged to combine the source-to-target and target-to-source model with the denoising auto-encoding and back-translation objectives. %We  mainly focus on the two steps in the following part.

\paratitle{Pre-training on Monolingual Corpora.}
Recent PLM-based research has mainly focused on the first step of UMT. Specifically, XLM~\cite{xlm} and mBERT~\cite{bert} were pre-trained on multiple monolingual data using MLM task, and then the PLM was used to initialize both the encoder and the decoder for machine translation. mBART~\cite{mbart} followed the pre-training scheme of BART~\cite{bart} on multiple languages, while these PLMs just performed the original pre-training task with mixed monolingual corpora, without considering the relationship between languages. CMLM~\cite{RenWLZM19} further proposed cross-lingual MLM to randomly mask tokens in monolingual sentences and predicted corresponding translation candidates. Therefore, CMLM was able to align the embeddings of different languages. CSP~\cite{csp} shared the similar idea, replacing some words in the source sentences with their translation words and then predicting the replaced words. 
%Moreover, creating pseudo-parallel corpora is also an effective way to augment monolingual datasets. MARGE~\cite{marge} retrieves a set of relevant texts in various languages and reconstructs the original text conditioned on the retrieved texts. 
% Keung~\etal\cite{KeungSLS20} propose to employ mBERT to calculate the sentence embedding of each sentence, and select the $k$ closest target sentences for each source sentence as pseudo-parallel corpora.

\paratitle{Leveraging Iterative Back-translation.}
In the back-translation stage, Garcia~\etal\cite{munmt} proposed using multi-task learning. They investigated \emph{multilingual UNMT}, which involved the use of a third language when translating one language into another. The extra language can provide auxiliary monolingual data or parallel data containing only one language in the source or target language. They aggregated back-translation loss and introduced a cross-translation term to incorporate the auxiliary corpus. Li~\etal\cite{runmt} also applied the cross-translation term and additionally included a knowledge distillation objective for the third (intermediate) language.
% Wang~\etal\cite{cunmt} optimize the loss in the unsupervised, supervised and cross-lingual direction.

\subsubsection{Supervised Machine Translation}
Supervised machine translation (SMT) refers to fine-tuning PLMs based on parallel corpora. Here, we will discuss how to utilize existing self-supervised PLMs and how to design PLMs for parallel corpora.

\paratitle{Directly Fine-tuning Unsupervised PLMs.} Almost all PLMs mentioned above using unsupervised (self-supervised) pre-training, such as XLM~\cite{xlm} and mBART~\cite{mbart}, can be directly fine-tuned with bilingual pairs. 
%Though pretrained on English-only corpora, BART is possible to be used for machine translation with a small additional encoder. The small encoder can be learned from bitext to map foreign words.
Moreover, considering the excellent encoding capability of BERT, BERT-fused model~\cite{bert_fuse} leveraged BERT to extract contextual embedding for the source sentence, and fused the representations with each layer of the encoder and decoder. CTNMT~\cite{ctnmt} leveraged asymptotic distillation and dynamic switching gate to integrate the BERT embedding.
%, and APT~\cite{apt} proposes dynamic fusion mechanism and knowledge distillation paradigm to learn and reverse the knowledge from BERT.
%Furthermore, 
Graformer~\cite{graformer} grafted mBERT as the encoder and mGPT as the decoder, and then trained a cross-attention module to combine them. Tang~\etal\cite{ml50} proposed to fine-tune mBART on multiple language pairs, which is called \emph{multilingual fine-tuning}.
%The majority of the above PLMs significantly improve zero-, low- or medium- resource translation, when compared with a randomly initialized Transformer~\cite{mbart}. These results demonstrate the effectiveness of pretraining on multilingual corpora. In contrast, PLMs, pretrained with unsupervised tasks, usually suffer from performance degradation in high-resource translation. In this case, multiple languages may reduce the weight capacity available for rich languages.

\paratitle{Designing PLMs for Parallel Corpora.} Most of PLMs are pre-trained on monolingual corpora using self-supervised pre-training tasks such as MLM and DAE. Nevertheless, these pre-training objectives are different from the downstream translation task. Hence, mRASP~\cite{mrasp} pre-trained the model on bilingual pairs with supervised Seq2Seq loss by randomly replacing the words in the source sentence with the words which have the same meaning in other languages. As a result, words with similar meaning across different languages are encouraged to share similar representations. 
% Similar to this idea, Yang~\etal\cite{alm} propose ALM to replace the source words with their target translations for better alignment.
mRASP2~\cite{PanWWL20} applied contrastive learning to minimize the representation gap of similar sentences and maximize that of %irrelevant 
unrelated sentences.
% Moreover, VECO~\cite{veco} proposes to plug a cross-attention module into the Transformer encoder part and designs a cross-attention MLM task to explicitly capture the interdependence between the representation of each language. During fine-tuning, the pretrained cross-attention modules can be the initialization of the decoder, which avoids initialize the cross-attention part randomly.
%Pretrained on parallel data, these models can improve machine translation for any pairs of language, including low-resource and high-resource languages. 
%In comparison to hundreds of billions of monolingual sentences, these models require only hundreds of million bilingual pairs, whereas the acquisition of annotated data requires massive manpower and financial resources.
Despite significant success, pre-training on parallel data requires massive labour and financial resources to create vast amounts of bilingual pairs.

\subsection{Text Summarization}
Text summarization is the process of condensing text into a brief summary that retains key information from the source text~\cite{summary_survey1}. The mainstream approaches to text summarization based on PLMs are either extractive or abstractive. Extractive summarization selects a subset of sentences from the source text and concatenates them to form the summary~\cite{LiuL19,hibert}. In contrast, abstractive summarization generates the summary automatically from the abstract representation of input texts~\cite{pointer,pegasus}. As abstractive summarization is more related to text generation, we only discuss  abstractive summarization in this section.

% According to the category of the textual contents, we will introduce document summarization, dialogue summarization and multilingual summarization in turn.

\subsubsection{Document Summarization}
Document is a widely-used literary form, such as news, opinions, reviews, and scientific papers.
% Considering the length of document, we categorize document summarization into vanilla summarization and long document summarization. Due to the length limitation of PLMs (usually $512$ or $1024$), vanilla summarization directly truncates the source document, while long document summarization commonly extends the architecture of PLMs for preserving more input information.
PLMs, such as UniLM~\cite{unilm,unilmv2}, MASS~\cite{mass}, T5~\cite{t5}, BART~\cite{bart} and PEGASUS~\cite{pegasus}, can be directly fine-tuned for document summarization. %PEGASUS~\cite{pegasus} is a PLM tailored for summarization. 
During pre-training, these models learn to predict the masked important sentences in the input document based on the remaining ones, which shares the similar idea of summarization. 
% Moreover, STEP~\cite{step} is pretrained to reinstate the original document, which can be seen as a sentence-level DAE pretraining.
%Thus, most of works utilized BART or PEGASUS as backbone for summarization.

Without directly generating summaries, several studies first extracted keywords, key sentences or relations as guidance and then combined these with PLMs for generation. CIT~\cite{cit} employed RoBERTa~\cite{roberta} to extract the important words and sentences from the input document.
%CTRLsum~\cite{ctrlsum}, similar to CIT, utilizes BERT to extract keywords  and also explores to control the entity and length of the summary using these keywords. 
In addition, topic models are used to capture the global topic semantics of the document, which can be integrated into the summarization model~\cite{NguyenLLQ21}. 
% FASum~\cite{fasum} extracts the factual relation triples and integrates them via graph attention.
GSum~\cite{gsum} proposed a general framework taking different kinds of guidance signals into the generation model, including keywords, triples, highlighted sentences and retrieved summaries. 
Apart from external guidance, several tricks can be applied to document summarization. 
Cao~\etal~\cite{CaoW21} improved the attention mechanism to emphasize salient content in the document. Refactor~\cite{refsum} first generated multiple summaries under different setups and then scored them and finally selected an optimal candidate summary. 

\subsubsection{Dialogue Summarization}
Dialogues, such as chat and medical conversation, consist of multi-turn utterances by two or more individuals. Hence, it is critical to capture the semi-structured dialogue content and users' interactions in dialogue~\cite{dialog_sum_survey}. 
For dialogue summarization, it is straightforward to direclty reuse document summarization models. 
%The method used in document summarization can be directly transferred into dialogue summarization.
Zhang~\etal~\cite{ZhangNGJHSG21} first truncated the dialogue text into several chunks, then summarized each chuck into partial summaries, and finally rewrote these partial summaries into a complete summary.

Meanwhile, several studies also explored some specific  
characteristics of dialogue for improving dialogue summarization. Chen~\etal~\cite{ChenY20} first extracted different topic views from conversations, and then utilized a multi-view decoder to combine these views for generating summaries. Furthermore, Chen~\etal~\cite{ChenY21} constructed discourse relation graphs and action graphs of conversations, in order to concentrate on the most salient utterances and understand concrete details of users' action. 
Considering the low information density, topic drifts and frequent coreferences of dialogue~\cite{dialog_sum_survey}, some researchers conducted auxiliary tasks to extract intrinsic information of dialogue. Feng~\etal~\cite{FengFQ0020} utilized DialoGPT~\cite{dialogpt}, a PLM specially designed for dialogue, to automatically extract keywords, detect redundant utterances and divide a dialogue into topically coherent segments. 

\subsection{Dialogue System}
Dialogue system (\aka conversational agent) aims to make machines communicate with human fluently. Technically, machines are required to generate a response conditioned on history contexts. According to downstream applications, dialogue systems are commonly categorized into open-domain and task-oriented dialogue systems. The former intends to converse with humans engaged on open topics such as daily life, sports and entertainment~\cite{odd}, while the latter is focused on assisting users to complete specific tasks, such as hotel reservation and product purchase~\cite{tod}. 

\subsubsection{Open-domain dialogue System}
Open-domain dialogue system is also known as chat-bots focusing on daily chat. For example, Microsoft XiaoIce is a well-known open-domain dialogue system to satisfy human needs for communication, affection, and social belonging~\cite{xiaoice}.

\paratitle{Continuous Pretraining with dialogue Corpora.}
PLMs, such as GPT-2, are pre-trained on general text corpora, thus various studies continually pre-trained general-purpose PLMs to fit dialogue systems. 
%extended typical PLMs to dialogue system by continuously pretraining them in dialogue corpora.
Due to the difficulty in obtaining large-scale dialogue corpora, informal text resources (such as forum posts and comments in Reddit, Twitter and Weibo) are usually employed for continual pre-training. 
As two typical models, 
DialoGPT~\cite{dialogpt} and Meena~\cite{meena} used English or Chinese dialogue corpora to continually pre-train casual LMs like GPT-2. Besides, Blender~\cite{blender} and PLATO~\cite{plato}  utilized the Seq2Seq loss to generate the next utterance based on previous utterances. 
Moreover, PLATO~\cite{plato} incorporated the next utterance classification (NUC) loss, similar to the next sentence prediction task in BERT, to judge whether the response is relevant to history dialogues to enhance the coherence of utterances. 
In order to penalize bland responses and decrease repetitions, DialoGPT~\cite{dialogpt} employed mutual information maximization to predict the input given generated response and Blender~\cite{blender} adopted unlikelihood training objective to penalize repetitive $n$-grams.
% In addition, Zhao~\etal~\cite{ZhaoXW20} propose four auxiliary tasks to pretrain including word order recovery, utterance order recovery, masked word recovery and masked utterance recovery to better understand dialogue and generate more relevant response. Zeng~\etal~\cite{ZengN21} compare different architectures of PLMs for dialogue system and propose solutions to alleviate discrepancies between pretraining and fine-tuning.

\paratitle{Directly Fine-tuning Existing PLMs.}
In addition to pre-training on dialogue corpora, researchers also explored fine-tuning existing PLMs on dialogue tasks. TransferTransfo~\cite{transfertransfo} adapted GPT to the dialogue task through multi-task learning. Based on TransferTransfo, Golovanov~\etal~\cite{GolovanovKNTTW19} modified the architecture to better model multiple inputs including dialogue history, persona information, and current state. 
% Mesgar~\etal~\cite{MesgarSG21} further enhance the factual consistency with dialogue history and persona information with reinforcement learning.
% BoB~\cite{bob} is composed of two decoders to better understand consistency of personas. 
% Santhanam~\cite{SanthanamCMDBHZ20} utilize two planners to generate plans for producing an informative response. ComPAC~\cite{compac} utilizes commonsense knowledge base or paraphrasing resources to expand persona information. 
Besides, to capture the hierarchical structure of dialogue, hierarchical encoders have been proposed to model the dialogue input~\cite{GuYH21,LiZF0Z20}. Gu~\etal~\cite{GuYH21} 
proposed a hierarchical framework, dialogueBERT, that uses sentence- and discourse-level Transformer encoders to encode each dialogue utterance and the sequence of utterance vectors, respectively. 
%dialogueBERT~\cite{GuYH21} employs a hierarchical Transformer architecture and additional training objectives to capture the discourse-level coherence of dialogue. DialoFlow~\cite{LiZF0Z20} proposes a dynamic flow mechanism to model the dialogue history by addressing the semantic influence of each utterance.
Furthermore, %various works focused on the controllable dialogue system.
controllability is also important to consider in dialogue systems.
Zeng~\etal~\cite{ZengN21} utilized condition-aware Transformer block to steer the response in a specific topic label. StyleDGPT~\cite{Yang0XLBWWL20} attempted to enforce the target style of 
the generated response  with KL loss at both word and sentence levels. 
% CoMAE~\cite{comae} introduces empathetic response generation considering user's communication mechanism, dialogue act and emotion in a hierarchical way. In order to generate coherent and informative response, some dialogues leverage external background, such as document~\cite{PrabhumoyeHZBS21} and knowledge~\cite{LiuZLRZY21,GaletzkaRS020}.

\subsubsection{Task-Oriented Dialogue System}
Task-oriented (\aka goal-oriented) dialogue system is a widely-used text generation application in real life, such as helping users order tickets.
% Before emergence of PLMs, task-oriented dialogue is typically broken down into several modules, including natural language understanding (NLU) to understand user's intent based on user's input, dialogue state tracking (DST) to track the constraints imposed by user and fulfill several slot-value pairs, dialogue policy learning (DPL) to determine next system action (\aka dialogue actions) involving an optional database state and natural language generation to generate according response based on system action in order to query more requirements~\cite{tod,simpletod}. 
Generally, task-oriented dialogue system was divided into four modules, \ie natural language understanding, dialogue state tracking, dialogue policy learning and natural language generation~\cite{tod}. 
%These components have labeled data to guide each module. 
%Though the main goal of task-oriented dialogue is to track user's intent and state to fulfill slot-value pairs, 

Most previous work only focused on the last generation module in task-oriented dialogue system by using generative PLMs (\eg GPT). For example, SC-GPT~\cite{PengZLLLZG20} used the ground-truth results of previous three modules (\eg dialogue state) and serialized them as input of the last generation module to generate response. Kale~\etal~\cite{KaleR20} further designed a manual schema to better convert previous results into a natural language. Shalyminov~\etal~\cite{Shalyminov20} proposed to  generate and retrieve several responses based on the dialogue context and utilized the NUC task to select the best one. PRAL~\cite{pral} utilized two separate GPT-2 to model the user and system, and adopted a third GPT-2 to perform knowledge distillation and incorporate commonsense knowledge into the final dialogue generation. Besides, more and more studies proposed to jointly learn these four modules based on a shared PLM. Budzianowski~\etal~\cite{BudzianowskiV19} and Hosseini-Asl~\etal~\cite{simpletod} generated the dialogue state, system action and final response successively, based on the original dialogue history.
% UBAR~\cite{ubar}, MinTL~\cite{mintl} and SOLOIST~\cite{soloist} share the similar idea that generating system response combining dialogue history, dialogue state, system action , database state or previous generated ones.
%In addition, 
%SOLOIST~\cite{soloist} employ the NUC loss just as open-domain dialogue system. 
%Shalyminov~\etal~\cite{Shalyminov20} proposed to generate and retrieve several candidate responses respectively and utilize the NUC task to select the best one. PRAL~\cite{pral} utilizes two GPTs to model user and system respectively, and also involves a third GPT to perform knowledge distillation.

\subsection{Others}
In this part, we will briefly introduce other text generation tasks, such as question generation, story generation and data-to-text generation. 

\subsubsection{Question Generation}
Question generation can be seen as a dual task of question answering (QA), \ie generate coherent questions based on given passages and answers. Existing PLMs, such as UniLM~\cite{unilm,unilmv2} and ProphetNet~\cite{prophetnet}, can be  employed for this task by taking as input the concatenation of the passage and answer. % XNLG~\cite{xnlg} and DeltaLM~\cite{deltalm} extend the question generation under multilingual scenario.
% Kulshreshtha~\etal~\cite{KulshreshthaBSR21} introduce back-training for unsupervised domain adaptation. SSR~\cite{ssr} proposes to refine the generated imperfect text into ground truth. And CRQDA~\cite{crqda} utilizes the rewriting method to conduct data augmentation.
Moreover, researchers explored this task in different QA settings. For example, Huang~\etal~\cite{HuangQSZ21} proposed a two-stage model to solve multi-hop question generation, and Cao~\etal~\cite{Cao020} attempted to generate open-ended questions which are answered by multiple sentences. 
Moreover, Majumder~\etal~\cite{MajumderRGM21} proposed a clarification question generation task to ask questions about the missing information in the passage in order to reduce the ambiguity. 
% Gu~\etal~\cite{chaincqg} explore the conversational question generation and Cheng~\etal~\cite{ChengLLZLLZ20} investigate the difficulty-controllable question generation. Besides, the conversational machine reading task also requires a question generation part to ask a follow-up question to trigger the next  turn~\cite{GaoWJXSKLH20,GaoWLJHXKL20,OuyangZZ21}

\subsubsection{Story Generation}
Story (or narrative, news) generation requires to generate a long-form open-ended text leveraging the given title or premise. It is challenging to produce a coherent and informative text based on limited input~\cite{tgsurvey}. To enrich the content of generated text, some studies aimed to incorporated external knowledge into PLMs. Guan~\etal~\cite{GuanHHZZ20} and Mao~\etal~\cite{MaoMMC19} utilized commonsense knowledge base to fine-tune PLMs to generate reasonable stories.
Megatron-Cntrl~\cite{megatroncntrl} used extracted keywords to retrieve knowledge sentences and then selected top-ranked sentences for story generation. 
%GRF~\cite{grf} conducts multi-hop reasoning on knowledge graph to provide concept distribution during generation. 
Besides, to generate coherent long-form text, 
%content planning is a common technique to organize the output structure.
%and then model generates output based on input and content plan (\aka outline).
PlotMachines~\etal~\cite{plotmachines} extracted keywords from input as outline to organize the output structure;
%ProGen~\cite{progen} iteratively refined the generated texts to enhance the quality. 
% Goldfarb-Tarrant~\etal~\cite{Goldfarb-Tarrant20} equip the planning model with an ensemble of rescoring models, each of which capture an element of good story-writing as laid out in Aristotle's Poetics. 
Guan~\etal~\cite{GuanHHZZ20} leveraged the contrastive learning loss to judge whether two sentences are consecutive in original text. 
% PermGen~\cite{permgen} maximize the likelihood of all possible output sentence orders. 
% HINT~\cite{hint} intermediate trains the PLM to predict inter-sentence similarity and the sentence order.
\ignore{
Besides the common story generation task, Plug-and-Blend~\cite{plugandblend} explore the generation with user-defined length and multiple topics. INSET~\cite{inset} and COINS~\cite{coins} investigate the story completion task, \ie infilling missing sentencing in a passage. Facts2Story~\cite{facts2story} generates the story conditioned on several facts (sentences). Qin~\etal~\cite{QinBHBCC19} propose the counterfactual story rewriting that rewrites the story to make it compatible with the given counterfactual event.}

\subsubsection{Data-to-text Generation}
The above tasks take unstructured text as input, while the data-to-text generation task generates descriptive text about structured input data, such as table, knowledge graph (KG) and abstract meaning representation (AMR). 
% Besides the mentioned forms, there also exists other input such as program code~\cite{pymt5} and math equation~\cite{0001LB21}. 
% Due to the data-hungry nature of data-to-text generation, PLMs can provide prior knowledge to boost the generation performance~\cite{ChenECLW20}. Besides, considering the descriptive nature of text, it is usual to employ pointer-network~\cite{pointer} to judge whether to directly copy the word in the structured input.
First, a naive and straightforward approach is
to   directly linearize the structured table~\cite{ChenECLW20,tablegpt} and KG~\cite{Ribeiro2020,HarkousGS20} into textual form as the input of PLMs. 
%Some researchers also designed \xin{special} pretraining tasks to pretrain \xin{specific} PLMs for table-to-text generation~\cite{XingW21}, KG-to-text generation~\cite{kgpt} and AMR-to-text generation~\cite{FanG20}. 
%In order to improve the controllability and faithfulness, Liu~\etal~\cite{0001ZCS21} and Su~\etal~\cite{SuVWFC21} utilize a planner to predict an ideal content plan.
Considering the graph structure of KG and AMR, Li~\etal~\cite{LiTZWYW21} and Ribeiro~\etal~\cite{RibeiroZG21} employed graph neural network to learn a better representation for each node. %And, Li~\etal~\cite{LiTZWYW21} further aligned the entity embedding of PLM and GNN to bridge the semantic gap.
Moreover, to cope with the structural information, a typical approach is to incorporate auxiliary training objectives such as predicting the value of table~\cite{tablegpt} and the relation of knowledge graph~\cite{LiTZWYW21}.

%\xin{introducing auxiliary tasks such as reconstructing the structured table~\cite{tablegpt} and KG~\cite{LiTZWYW21,KeJRCWSZH21} is usually used to capture the semantic correspondence between structured input and output text. (entirely rewriting)} 
%Some studies borrowed the idea of dual learning to jointly learn the data-to-text generation and text-to-data parsing tasks~\cite{KeJRCWSZH21}. 
%Su~\etal~\cite{SuMBC21} and Chang~\etal~\cite{ChangSZDS21} retrieve additional information to argument the input data. And Suadaa~\etal~\cite{SuadaaKFOT20} and Chen~\etal~\cite{ChenCSCW20} propose a new variant of table-to-text generation involving logical reasoning.

\subsubsection{Other Generation Tasks}
Besides the aforementioned tasks, there are also other text generation applications. ColdGANs~\cite{coldgans} explored the unconditional language generation. KG-BART~\cite{kgbart} investigates the commonsense generation, \ie generating a natural language consisting of provided commonsense concept (word), which can be considered as the hard-constrained conditional generation~\cite{tgsurvey}. Moreover, text style transfer aims to convert a text into another style while preserving the basic semantics of input~\cite{tgsurvey}, such as sentiment transfer and writing style transfer~\cite{strap}. In addition, some researchers devoted to literary creation, such as poem~\cite{songnet} and lyric~\cite{deeprapper}.
\section{Conclusion and Future Directions}
\label{sec:con}

In this survey, we presented an overview of current representative research efforts on PLMs-based text generation, and expect it can facilitate future research. We began with introducing three key aspects when applying PLMs to text generation, based on which the main content of our survey is divided into three sections from the view of input representation learning, model architecture design, and parameter optimization. Besides, we discussed several non-trivial challenges related to the above three aspects. Finally, we reviewed various evaluation metrics, open-source libraries, and common applications to help practitioners evaluate, choose and employ PLMs for text generation. 

Despite the great progress made in recent years, we are faced with %To advance this field, there remains 
several open problems and several future directions are promising to deal with them.

\paratitle{Controllable Generation.} Controllable text generation with PLMs is an interesting direction but still at a very early stage. Controlling some attributes of the generated text has many practical use cases, such as generating positive responses to patients suffering from depression in dialogue systems. However, PLMs are usually pre-trained in universal corpora, which is difficult to control the multi-grained attributes of the generated text (\eg sentiment, topic, and coherence). \citet{ctrl} has explored text generation with control codes that govern style, content and task-specific behavior. However, these control codes are preset and coarse-grained. Future work can explore multi-grained control and develop PLMs that are sufficiently steerable.

\paratitle{Optimization Exploration.} Fine-tuning is the predominant optimization way to distill the linguistic knowledge stored in PLMs to downstream generation tasks. Now, prompt-based learning has become a performant and lightweight optimization method~\cite{prompt-survey}. Future work can explore a broader range  %more kinds 
of optimization approaches that can combine the advantages of current methods.

%. Hence, we advocate for research on more transfer methods of PLMs for text generation.

\paratitle{Language-agnostic PLMs.} Nowadays,  almost all the PLMs for text generation are mainly for English. 
These PLMs will encounter challenges when dealing with non-English generation tasks. 
%However, English is not the native language for the majority of the world's population, which becomes an obstacle to non-English text generation.
Therefore, language-agnostic PLMs are worthy to be investigated. This requires us to capture universal linguistic and semantic features across different languages. An interesting direction is explore how to reuse existing English-based PLMs for text generation in non-English languages.%Rather than cross-lingual translation, it is more convenient to add simple modules to reuse current PLMs for generation.

\paratitle{Ethical Concern.} Currently, PLMs are pre-trained on large-scale corpora crawled from web without fine-grained filtering, potentially causing  ethical issues such as generating private content about users. Therefore,  researchers should try their best to prevent misusing PLMs. 
%Current PLMs can generate text realistic enough and therefore it has potential to result in harmful applications. 
%Although it is somewhat difficult to anticipate misuse of PLMs, we should try best to prevent it. 
%For this purpose, we can follow the key steps in~ \cite{guide4conductrisk}, such as identifying threats and potential impacts and assessing likelihood. 
Besides, the text generated by PLMs might be prejudiced, which is in line with the bias in training data along the dimensions of gender, race, and religion~\cite{gpt3}. As a result, we should intervene PLMs for preventing such biases. The research on the general approach is extensive but still preliminary for PLMs.

In conclusion, text generation based on PLMs has greatly contributed to the advance of the state of the art in this field. However, the current state of the art in different text generation tasks is still far from what one could expect. Extensive research efforts are needed to better adapt PLMs to text generation tasks.

\bibliographystyle{ACM-Reference-Format}
\bibliography{survey}


\end{document}